\documentclass[]{fairmeta}
\usepackage{amssymb}
\usepackage{amsmath}
\usepackage{multirow}
\usepackage{todonotes}
\usepackage{wrapfig}
\usepackage{url}
\usepackage{biblatex}
\usepackage{xcolor}
\usepackage{dirtree}
\usepackage{graphicx}
\usepackage{float}

\usepackage{listings}
\usepackage{xcolor}

\definecolor{codebg}{RGB}{245,245,245}
\definecolor{codekey}{RGB}{0,0,180}       
\definecolor{codestr}{RGB}{163,21,21}     
\definecolor{codecom}{RGB}{0,128,0}       
\definecolor{codenum}{RGB}{128,0,128}     
\definecolor{codeframe}{RGB}{200,200,200}

\lstdefinestyle{codebase}{
  basicstyle=\ttfamily\small,
  backgroundcolor=\color{codebg},
  frame=single,
  rulecolor=\color{codeframe},
  framesep=4pt,
  breaklines=true,
  breakatwhitespace=true,
  columns=fullflexible,
  keepspaces=true,
  showstringspaces=false,
  tabsize=2,
  captionpos=b,
  keywordstyle=\color{codekey}\bfseries,
  stringstyle=\color{codestr},
  commentstyle=\color{codecom}\itshape,
}

\lstdefinelanguage{json}{
  morestring=[b]",
  morestring=[d]',
  stringstyle=\color{codestr},
  literate=
    *{:}{{{\color{codekey}{:}}}}{1}
     {,}{{{\color{codekey}{,}}}}{1}
     {\{}{{{\color{codekey}{\{}}}}{1}
     {\}}{{{\color{codekey}{\}}}}}{1}
     {[}{{{\color{codekey}{[}}}}{1}
     {]}{{{\color{codekey}{]}}}}{1},
}

\lstdefinelanguage{yaml}{
  keywords={true,false,null,True,False,Null,yes,no,on,off},
  keywordstyle=\color{codekey}\bfseries,
  sensitive=true,
  comment=[l]{\#},
  commentstyle=\color{codecom}\itshape,
  morestring=[b]',
  morestring=[b]",
  stringstyle=\color{codestr},
  literate=
    *{:}{{{\color{codekey}{:}}}}{1}
     {-}{{{\color{codekey}{-}}}}{1},
}

\makeatletter
\newcommand\footnoteref[1]{\protected@xdef\@thefnmark{\ref{#1}}\@footnotemark}
\makeatother

\urldef{\foundryWebsite}\url{https://tri-ml.github.io/vla_foundry}
\urldef{\foundryCodebase}\url{https://github.com/TRI-ML/vla_foundry}
\urldef{\hfmodels}\url{https://huggingface.co/collections/TRI-ML/vla-foundry}

\newcommand{\lbm}{\textsc{LBM}}
\newcommand{\lbmmt}{\textsc{LBM-MT}}
\newcommand{\foundryLLM}{\textsc{Foundry-LLM-1.2B}}
\newcommand{\foundryVLM}{\textsc{Foundry-VLM-1.3B}}
\newcommand{\foundryVLA}{\textsc{Foundry-VLA-1.7B}}
\newcommand{\foundryQwenVLA}{\textsc{Foundry-Qwen3VLA-2.1B-MT}}
\newcommand{\foundryVLASim}{\textsc{Foundry-VLA-1.7B-MT-sim}}
\newcommand{\foundryVLAReal}{\textsc{Foundry-VLA-1.7B-MT-real}}
\newcommand{\foundryVLAMT}{\textsc{Foundry-VLA-1.7B-MT}}

\newcommand{\lbmevalcs}{\texttt{lbm\_eval\_cs}}
\newcommand{\lbmevaloss}{\texttt{lbm\_eval\_oss}}

\title{VLA Foundry: A Unified Framework for Training Vision-Language-Action Models}

\author[*\dagger]{Jean Mercat}
\author[*\dagger]{Sedrick Keh}
\author[\dagger]{Kushal Arora}
\author[\dagger]{Isabella Huang}
\author[\dagger]{Paarth Shah}
\author[]{Haruki Nishimura}
\author[]{Shun Iwase}
\author[\dagger]{Katherine Liu}

\affiliation[]{Toyota Research Institute}

\contribution[*]{Co-first authors}
\contribution[\dagger]{Core contributors}

\abstract{%

\vspace{20pt}
We present VLA Foundry, an open-source framework that unifies LLM, VLM, and VLA training in a single codebase. Most open-source VLA efforts specialize on the action training stage, often stitching together incompatible pretraining pipelines. VLA Foundry instead provides a shared training stack with end-to-end control, from language pretraining to action-expert fine-tuning. VLA Foundry supports both from-scratch training and pretrained backbones from Hugging Face. To demonstrate the utility of our framework, we train and release two types of models: the first trained fully from scratch through our LLM$\rightarrow$VLM$\rightarrow$VLA pipeline and the second built on the pretrained Qwen3-VL~\cite{bai2025qwen3} backbone. We evaluate closed-loop policy performance of both models on LBM Eval~\cite{lbm_eval2025}, an open-data, open-source simulator. We also contribute usability improvements to the simulator and the STEP~\cite{snyder2025step} analysis tools for easier public use. 
In the nominal evaluation setting, our fully-open from-scratch model is on par with our prior closed-source work ~\cite{lbmtri2025} and substituting in the Qwen3-VL backbone leads to a strong multi-task table top manipulation policy outperforming our baseline by a wide margin.

The VLA Foundry codebase is available at \foundryCodebase{} and all multi-task model weights are released on \hfmodels{}. Additional qualitative videos are available on the project website \foundryWebsite{}.
\vspace{20pt}
}
\date{\today}
\correspondence{Jean Mercat and Katherine Liu at \email{firstname.lastname@tri.global}}

\metadata[Project page]{\url{https://tri-ml.github.io/vla_foundry/}}
\begin{document}
\maketitle
\newpage

\section{Introduction}
\label{sec:intro}

Robotics foundation models are advancing at a rapid pace, with many systems~\cite{black2025pi05, intelligence2025pi06star, shukor2025smolvla, molmoact2025, dreamzero_ye2026worldactionmodelszeroshot, lin2026holobrain} demonstrating capabilities that would have seemed out of reach just a few years ago. As the frontier moves faster, the tooling required to support rigorous research must keep pace. 
Many high-impact questions -- about data scaling, backbone pretraining, and the interplay between robotics and non-robotics data -- require both scale (compute, data, etc.), as well as modular algorithmic infrastructure that allow users full control over different parts of the model and training pipeline.
However, most existing codebases have either not been extensively tested at scale~\cite{galahad2025vlascratch}, or are largely focused on model releases ~\cite{molmoact2025, physicalintelligence2025openpi, gr00tn1_2025} and therefore tightly coupled to specific algorithmic decisions, limiting research flexibility.

At the same time, data scarcity remains a fundamental bottleneck in robotics. Robot interaction data is severely constrained relative to data used for language and vision models, especially in diversity and in signal density per token~\cite{bandaru2025foundation}. As robot policies continue to scale, the relative importance of non-robotics data only grows~\cite{lin2026systematic}. Despite this data disparity, most open-source VLA frameworks focus narrowly on the action training stage, treating the upstream data recipe as fixed or out-of-scope. Such separation is problematic: data decisions made during LLM and VLM pretraining have direct consequences for downstream robotics performance. Exploring the design space requires a framework that treats the entire pipeline, from pretraining to policy learning, as a single, controllable system.

We developed \textbf{VLA Foundry} to address these challenges. VLA Foundry is a unified, open-source framework with a shared data-loading and training stack that spans LLM, VLM, and VLA training in a single codebase, giving practitioners control over the entire pipeline -- from backbone pretraining to action-expert fine-tuning. Because every stage shares the same infrastructure, researchers can co-train across modalities, mix datasets, and prototype new architectures without stitching together disparate tools. The framework natively supports pretrained backbones from Hugging Face, and its modular, configuration-driven design lets users swap architectures, data pipelines, and training recipes through simple command-line or YAML changes.

VLA Foundry has the following key features:
\begin{itemize}
\item \textbf{Full pipeline controllability}, enabling researchers to intervene at any
stage of the data and training recipe — from backbone pretraining to action expert
fine-tuning — through a shared, configuration-driven interface.

\item \textbf{Flexible multi-modal training} with probabilistic dataset mixing and dataloaders that support text, image-caption, and robotics data, allowing precise control over the training mixture at every stage.

\item \textbf{Native Hugging Face integration} facilitates loading of pretrained vision encoder, LLM, and VLM backbones, making benchmarking new architectures straightforward within the same training and evaluation pipeline.

\item \textbf{Scalable distributed training} built on FSDP2 and cloud-native tooling
(AWS SageMaker, S3), supporting multi-node, multi-GPU runs with automatic gradient
accumulation, mixed precision, and checkpoint synchronization.

\end{itemize}

\begin{figure}[h]
\centering
    \includegraphics[width=1\linewidth, trim={4.5cm 4cm 4.5cm 4cm}, clip]{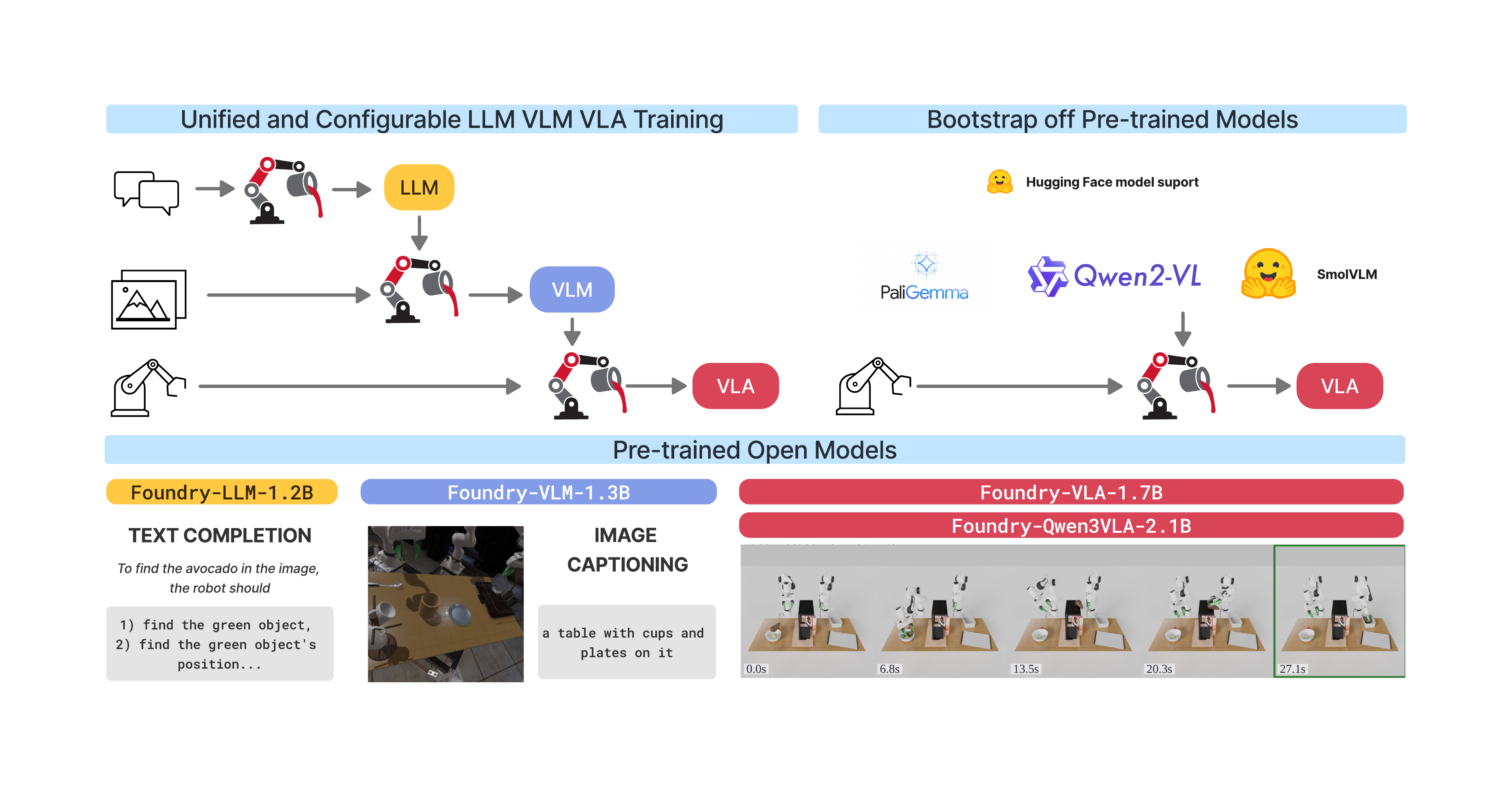}
    \caption{VLA Foundry overview. \textit{Unified and Configurable LLM-VLM-VLA Pipeline}: VLA Foundry was designed to enable flexible model composition. For example, users can train an LLM, use the LLM to train a VLM, and use the VLM to train a VLA. \textit{Bootstrap off Pre-trained Models}: VLA Foundry also supports loading off-the-shelf VLMs from Hugging Face. \textit{Pre-trained Open Models}: We release LLM, VLM, and VLA models trained both from scratch and finetuned from open weights under a permissive license at \hfmodels{}.}
    \label{fig:fig1}
\end{figure}
\section{Related Work}
\label{sec:related}

\subsection{LLM/VLM Training Frameworks}

Large Language Models (LLMs) form the foundation of modern multimodal systems, providing scalable sequence modeling capabilities, strong linguistic representations, and emergent reasoning abilities. Early work established the effectiveness of the scaling of transformer-based language models while subsequent efforts have largely focused on improving training efficiency, transparency, and reproducibility. Projects such as Megatron-LM \cite{megatron-lm}, DeepSpeed~\cite{rasley2020deepspeed}, and GPT Neox~\cite{gpt-neox-library} introduced distributed training strategies that enabled scaling to hundreds of billions of parameters. More recent and accessible open training initiatives, including OpenLM~\cite{open_lm}, Olmo \cite{olmo20242olmo2furious}, LLM360~\cite{liu2023llm360}, and its follow-up K2 model~\cite{liu2025k2} emphasize full stack transparency by releasing training data, code, intermediate checkpoints, and logs. Complementary efforts such as FastLLM~\cite{fastllm2024} provide practical recipes for training competitive models under more constrained compute budgets, while educational repositories such as nanoGPT~\cite{karpathy2022nanogpt} and LLMs from scratch~\cite{build-llms-from-scratch-book} have further lowered the barrier to reproducing end-end-end LLM training pipelines. Dataset frameworks such as DCLM~\cite{li2024datacomp} and FineWeb~\cite{penedo2024fineweb} provide high quality language datasets.  Together, these works highlight a shift towards reproducible and modular LLM training frameworks.

Vision-language model (VLM) frameworks must additionally address cross-modal representation learning and heterogeneous data integration. A common and prominent design pattern in VLMs is to couple a pretrained vision encoder with language model backbone, with intermediate modules responsible for aligning visual and text representations. Frameworks such as OpenFlamingo~\cite{awadalla2023openflamingo} operationalize this approach by providing infrastructure for interleaved image-text sequence construction, multimodal batching, and various architecture choices that enable the training of autoregressive VLMs on web-scale data. Similarly, LLaVA~\cite{liu2023visual} offers a streamlined pipeline for multimodal instruction tuning, including data formatting, visual feature extraction, and supervised fine-tuning stages.

Other frameworks emphasize modularity and composability as first-class design principles. BLIP-2~\cite{li2023blip2} introduces a modular bridging component (Q-former) that decouples vision and language backbones, allowing each to be reused independently. Prismatic VLMs~\cite{karamcheti2024prismatic} extend this idea by explicitly structuring the framework around interchangeable components, enabling controlled experimentation over vision encoders, language models, and training mixtures. InternVL~\cite{chen2024internvl} further demonstrates how such modular designs can be scaled, incorporating large vision encoders and staged alignment strategies within a unified pipeline. Qwen~\cite{Qwen-VL, bai2025qwen3} offers state-of-the-art VLM capabilities in an open-source codebase. Complementing this line of work that primarily focuses on model architecture, dataset frameworks such as DataComp~\cite{gadre2023datacomp} provide standardized pipelines for construction and evaluating large-scale image-text datasets, addressing a critical bottleneck in reproducible multimodal training. 

Across LLM and VLM settings, these frameworks expose several key dimensions of design including  data pipelines (e.g. pre-tokenized vs. dynamic processing), model composition (e.g. monolithic vs. modular architectures), and training orchestration (e.g. distribution execution, staged vs joint optimization). As a result, existing systems span a spectrum from highly optimized distributed training backends to more modular, research oriented frameworks that facilitate experimentation with model architecture and data mixtures. 

\subsection{VLA Training Frameworks}

In recent years, the open source vision-language-action (VLA) ecosystem has expanded rapidly, moving from a small number of isolated model releases to a broader set of training pipelines, pre-trained checkpoints and reproducible research frameworks. One of the earliest milestones of this shift was OpenVLA \cite{kim2024openvla} which released a 7B-parameter model with a full PyTorch compatible codebase built off Prismatic \cite{karamcheti2024prismatic}. Since then, a number of open-source alternatives have emerged. OpenPi \cite{physicalintelligence2025openpi} provides training and fine-tuning support for Physical Intelligence's $\pi_0$ model series, with base checkpoints pretrained on more than 10,000 hours of robot data. GR00T \cite{gr00tn1_2025} pairs a vision-language backbone with a diffusion transformer action head in a dual-system architecture trained on real, simulated, and synthetic data. MolmoAct \cite{molmoact2025} explores a complementary direction by introducing an ``Action Reasoning Model'' that reasons in 3D space via depth-aware perception tokens rather than purely language-based action representations.

Beyond individual model development, several efforts have focused on standardization, infrastructure, and reproducibility of the entire VLA pipeline. LeRobot \cite{cadene2024lerobot} adopts a community-first approach, emphasizing affordable hardware (SO-100/101 arms), integrating dataset collection, training, and deployment across affordable hardware platforms and lowering the barrier to real-world experimentation. They report results on a 450M-parameter model, SmolVLA \cite{shukor2025smolvla}, which is trained on a single GPU and remains competitive with much larger VLAs on standard benchmarks. VLAb \cite{aubakirova2025vlab} complements LeRobot as a dedicated pretraining library and SmolVLA reproduction kit. VLA-Scratch \cite{weng2026vlascratch} provides a modular, performance-oriented training stack built on PyTorch FSDP2 with support for multiple VLM backbones (Qwen3-VL, PaliGemma, SmolVLM), heterogeneous dataset co-training, and Hydra-based configuration for rapid experimentation.
StarVLA \cite{community2026starvla} further advances this direction by explicitly decoupling backbone architectures from action heads and supports both VLM backbones (e.g., Qwen-VL) and world-model backbones (e.g., Cosmos) with multiple options for action heads (autoregressive tokens, continuous decoding, and flow-matching), and integrates multiple benchmarks through a unified evaluation interface. Dexbotic \cite{xie2025dexbotic} takes an experiment-centric approach, adopting a unified PyTorch toolbox with optimized reimplementations of various VLAs across different platforms such as the Franka and SO-101. 

\section{VLA Foundry}
\label{sec:foundry}

\textbf{VLA Foundry} is an open-source framework for training LLMs, VLMs, and VLAs within a single codebase. It is designed around end-to-end control of the embodied-model pipeline: the same training loop, data abstractions, and configuration interface extend from language pretraining to vision-language training and action learning. In this sense, VLA Foundry connects capabilities often treated separately across LLM/VLM training frameworks~\cite{megatron-lm, olmo20242olmo2furious, karamcheti2024prismatic} and VLA frameworks~\cite{kim2024openvla, physicalintelligence2025openpi, gr00tn1_2025, cadene2024lerobot, weng2026vlascratch, community2026starvla}. For robotics, this unified stack makes it practical to build and scale VLA systems while exploring new training recipes, architectures, and data mixtures. It supports both pre-training from scratch or initialization from pretrained Hugging Face backbones without requiring users to switch codebases across stages. An accompanying tutorial illustrates the full LLM$\rightarrow$VLM$\rightarrow$VLA training path from scratch\footnote{\url{https://github.com/TRI-ML/vla_foundry/tutorials/training_llm_vlm_vla.ipynb}}.

In this section we present the key elements of the VLA Foundry framework that we believe make it a useful tool for policy pretraining research and experimentation.

\subsection{Design Principles}
\label{sec:design-principles}
VLA Foundry is designed around four principles. We state them here; Section~\ref{sec:architecture} shows how the architecture embodies each, and Appendix~\ref{app:foundry-details} gives the full reference.

\begin{enumerate}
    \item \textbf{Modularity and Composability} -- Components plug together rather than being baked into the training code. Models, data pipelines, encoders, and loss handlers are instantiated by name from a YAML-based configuration system that supports nested includes, so presets can be composed, locally overridden, and reused across experiments; swapping a vision encoder, a language backbone, or an entire model type is a single command-line change.
    \item \textbf{Hackability and Interoperability} -- Any component can be extended or replaced without touching the rest of the system. We avoid heavy framework wrappers (PyTorch Lightning, Hugging Face Trainer) and keep the training loop thin with parallelism primitives exposed rather than hidden, so users are not locked into a particular stack and can extend the framework with new modeling architectures or distributed-training paradigms as they emerge.
    \item \textbf{Performance} -- VLA Foundry targets researchers with moderate to medium-scale compute. Training throughput has been benchmarked across LLM, VLM, and VLA stages up to 128 GPUs across 16 nodes.
    \item \textbf{Reproducibility} -- Runs are repeatable at a given configuration. We rely on deterministic per-rank RNG seeding, dataloader state checkpointing for exact restarts, and immutable frozen dataclasses that prevent hidden configuration changes at runtime.
\end{enumerate}

\subsection{Framework}
\label{sec:architecture}
VLA Foundry's architecture has four layers: a YAML-based configuration system backed by frozen dataclasses, a registry that makes models and data pipelines pluggable, modality-specific preprocessing and dataloading, and a model-agnostic training loop. The remainder of this section walks through each; Appendix~\ref{app:framework} gives the full reference.

\subsubsection{Modular Configuration System}
VLA Foundry's modularity and composability is ensured by our configuration system. We base it on Draccus~\cite{draccus}: every parameter is declared in a dataclass and can be overridden by a YAML preset or a command-line argument, in increasing order of priority. Presets themselves are composable -- a YAML file can inherit from others, so experiments are expressed by combining building blocks rather than by duplicating them. Parameters shared across modules (e.g., hidden dimensions, sequence lengths) are resolved once and propagated through the dataclass tree preventing silent cross-module mismatches. Configuration dataclasses are frozen to avoid run-time configuration changes that easily result in discrepancies between configuration files, logs, and runtime. See Appendix~\ref{app:config-system} for details and a worked example.

\subsubsection{Extending the Framework}
VLA Foundry is designed to be hackable and extensible. Adding a model that fits an existing model type (LLM, VLM, or DP-VLA) requires only a parameter dataclass and a factory function, registered by name at import time; the model type's \emph{batch handler} -- which owns batching, loss construction, and output reduction -- is shared, and a single training loop drives all model types. A new batch handler is needed only when introducing a new training paradigm.
Adding a dataset follows a similar pattern. Raw data is converted to WebDataset~\cite{breuel2020webdataset} tar shards through a per-modality preprocessing stage. Preprocessing runs in parallel with Ray~\cite{moritz2018ray} and emits both training shards and the per-dataset statistics needed for normalization at training time.
The dataloading pipeline itself is an ordered composition of small stages that users extend or reorder independently from the training loop. Dataloader can be mixed and each dataset contributes its own shards, statistics, and modalities with weighted dataset proportions.

\subsubsection{Robotics Data Handling}
Robotics data carries structure beyond what text and image-caption pipelines handle. Normalization is known to require careful handling in multi-dataset robotics training~\cite{lbmtri2025}; our \texttt{RoboticsNormalizer} supports global and per-timestep schemes, including percentile-based variants. Statistics can be merged across datasets. For percentile estimation and merging, we use t-digest~\cite{dunning2019computing}. Actions may be represented in absolute world-frame coordinates or relative to an anchor end effector pose, with rotations in the 6D continuous format~\cite{zhou2019continuity} and relative poses composed in SE(3). Actions are chunked~\cite{Zhao-RSS-23} in a configurable window of past and future time steps around an anchor: the future portion supervises the model, the past portion is available as input. Proprioceptive observations are causally restricted to past and current time steps. See Appendix~\ref{app:robotics}.

\subsubsection{Training Performance}
\label{sec:throughput}

The training loop supports the standard levers for scaling distributed training -- FSDP (with optional CPU offloading), mixed precision, gradient checkpointing, \texttt{torch.compile}, and gradient accumulation. See Appendix~\ref{app:model-training}.
Figure~\ref{fig:scaling plots} shows the training throughput across the different stages of our pipeline (LLM, VLM, and VLA). We used a 1.2 billion parameter language model, add a 86 million parameter ViT for the VLM and add a 325 million parameter transformer action head for the VLA. For the LLM, we used a sequence length of 2048 tokens with padding if needed. For the VLM, each image is represented with 64 tokens and the caption inputs are variable lengths but for consistency, we chose a total length of 256 tokens, truncated and padded sequences as needed. For the VLA, the model encodes 8 images from different cameras and timesteps, producing 512 tokens and a short task description. We pad VLA sequences dynamically and the average sequence length is 549 tokens. At this model scale, each GPU can hold the full model weights during training thus FSDP doesn't offer an advantage, and even shows weaker scaling for the VLM.

\begin{figure}[H]
    \centering
    \includegraphics[width=1\linewidth]{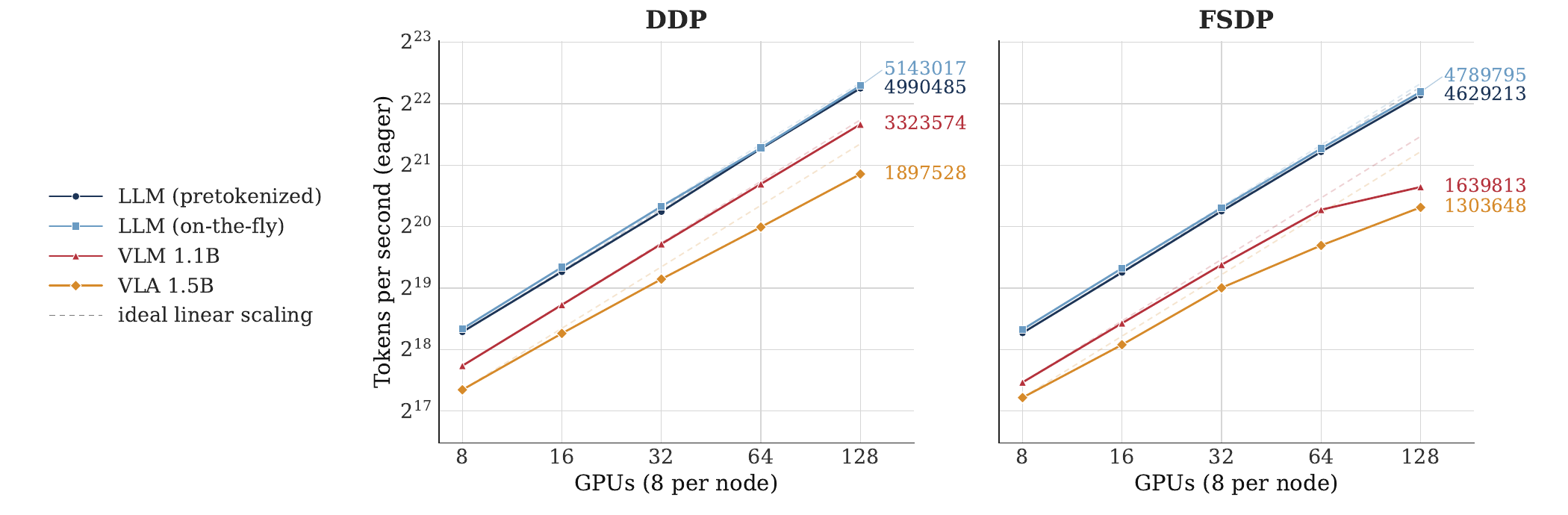}
    \caption{Throughput scaling as the number of GPU nodes is increased for the LLM, VLM, and VLA with either DDP or FSDP parallelization. Tests were done through Sagemaker on P5 nodes of 8 Nvidia H100 GPUs each. See section~\ref{sec:models} for details about the models.}
    \label{fig:scaling plots}
\end{figure}

\subsection{Evaluation}
\label{sec:dashboard}
VLA Foundry supports evaluation on \lbmevaloss{}, the open-source release of the \lbm{} simulation benchmark~\cite{lbmtri2025}. The \lbmevaloss{} framework is a challenging benchmark that uses the high fidelity Drake physics engine \cite{drake} to model the robots and scene dynamics. It defines 49 tasks to measure the performance of table top bimanual manipulation policies. Users can compare their own trained policies against the released checkpoints under a shared protocol. We ship the simulator as a Docker image, sidestepping platform-specific build and dependency issues across user environments. A simple dashboard lets users manage evaluation experiments, view rollout videos, and plot results as they accumulate.

Additionally, we provide rigorous statistical analysis via STEP~\cite{snyder2025step} to compare success rates of multiple policies. Following~\cite{lbmtri2025, lin2026systematic, lerobot_unfolding_robotics}, the dashboard has violin plots for Bayesian estimates of individual success rates, with Compact Letter Display (CLD)~\cite{piepho2004algorithm} attached for comparison. Policies not sharing any CLD alphabet are significantly different at 5\% family-wise error rate (FWER). Notably, our statistical framework lets the user base decisions on intermediate comparisons, as results are gathered. The user can decide to stop an evaluation early to save time, or to collect more rollouts than initially planned to seek higher statistical power. Such a practice would constitute harmful ``p-hacking''~\cite{stefan2023big} for standard statistical tests such as Barnard's test~\cite{barnard_significance_1947}. More details can be found in prior work~\cite{snyder2025step, tri_2026_ab_testing, lin2026systematic}; we include our general design principles and suggested best practices as documentation in the dashboard. In particular, when concatenating results over multiple tasks for aggregate comparison, we balance the per-task sample size for each policy to ensure that the aggregate represents an unbiased estimate of the policy's equally-weighted multi-task performance. For instance, if some Model A has $[50, 49, 50, 50]$ rollouts across 4 tasks, where the second task is missing one rollout, the results are truncated to $[49, 49, 49, 49]$ before aggregation. Therefore, Model A's aggregated performance is computed by 196 rollouts instead of 199 before it is fed to STEP for comparison. We note that this unbiased aggregation was not strictly enforced in the prior work~\cite{lbmtri2025}. Our results as well as those from~\cite{lbmtri2025} are included in the dashboard so that users can compare their own experiments with the reported numbers from the released checkpoints.

\section{\foundryVLA{} and \foundryQwenVLA{}}
\label{sec:models}

Having described the framework itself, we now turn to two applications. We release two VLA models types alongside this report. Each showcase different capabilities of the VLA Foundry pipeline:

\begin{itemize}
    \item \textbf{\foundryVLA{}} -- trained fully from scratch along the LLM$\rightarrow$VLM$\rightarrow$VLA pipeline, demonstrating end-to-end controllability over the training recipe.
    \item \textbf{\foundryQwenVLA{}} -- trained on top of a pretrained Qwen3-VL 2B backbone, showing that the same codebase efficiently supports the traditional VLM$\rightarrow$VLA recipe and that a stronger/larger VLM backbone translates into a more capable VLA.
\end{itemize}

Both models share the same action expert architecture (Section~\ref{sec:vla-data-and-arch}). Section~\ref{sec:vlaf-v0} walks through the from-scratch pipeline, Section~\ref{sec:vlaf-qwen3} describes the Qwen3-based model, and Section~\ref{sec:sim-eval} reports simulation results, including ablations over multi-task vs.\ single-task training as well as sim-only and real-only subsets.

\subsection{\foundryVLA{}: Training From Scratch}
\label{sec:vlaf-v0}

\foundryVLA{} is our end-to-end demonstration of VLA Foundry's full-pipeline controllability. We first train a language model (LLM), extend it to a vision-language model (VLM), and finally adapt it into a vision-language-action (VLA) model (Figure~\ref{fig:fig1}). We release the intermediate \foundryLLM{} and \foundryVLM{} checkpoints in addition to \foundryVLA{} so that the community can reproduce or modify any stage of the pipeline\footnote{\hfmodels{} \label{hfmodel}}.

\paragraph{\textbf{LLM training}} We used a standard transformer architecture~\cite{open_lm} to define a 1.2 billion parameter model with a hidden dimension of 2048, 24 layers, and 16 heads. Note that, following the convention~\cite{kaplan2020scaling}, we discount the additional 200 million parameters of the embedding layers.

The model was trained on 500 million samples (or 1 trillion tokens) from the openly available DCLM~\cite{li2024datacomp} dataset with a sequence length of 2048. Text was tokenized with the processor \texttt{HuggingFaceTB/SmolVLM2-256M-Video-Instruct}, which has a vocabulary size of 49,280.
We used a warmup-stable-decay learning rate schedule~\cite{hu2024minicpm}. The model and its full set of configuration parameters is available on HuggingFace\footnoteref{hfmodel}. Table~\ref{tab:llm_benchmarks} shows results of this model on standard benchmarks before the learning rate decay phase and after the full training. Note the lack of instruction tuning and the size of the model keep it close to random chance on difficult benchmarks such as MMLU; however, we see good results well above random chance on easier benchmarks.

\begin{figure}[h]
  \centering
  \begin{tabular}{cccc}
    \includegraphics[width=0.2\linewidth]{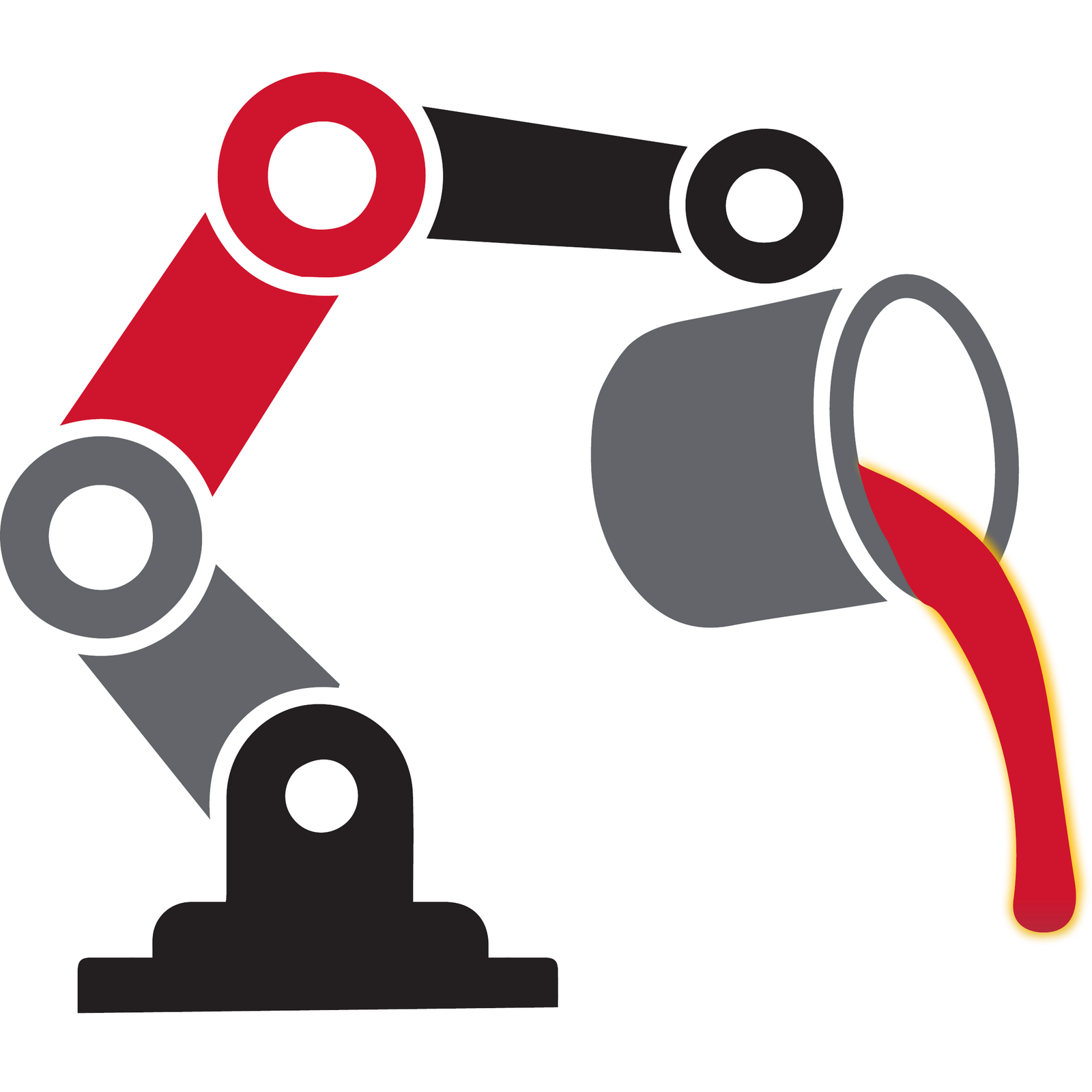} &
  \includegraphics[width=0.2\linewidth]{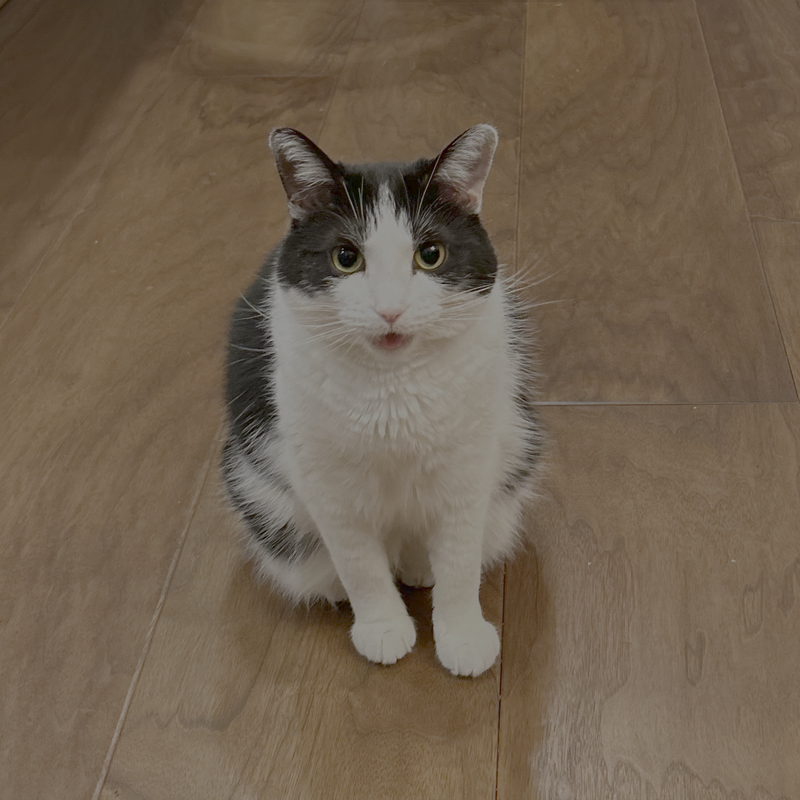} &
  \includegraphics[width=0.2\linewidth]{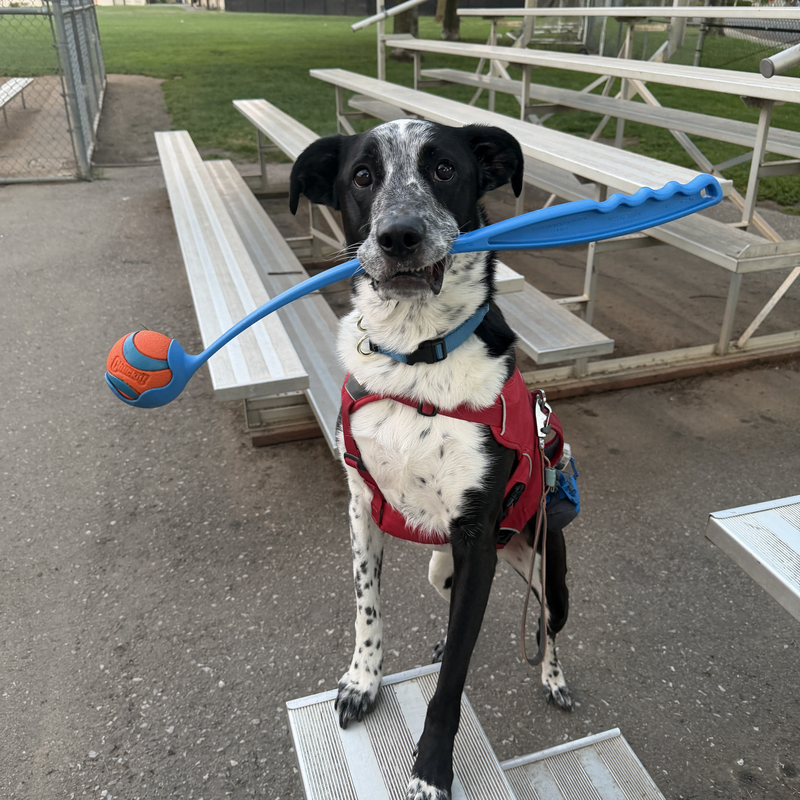} &
  \includegraphics[width=0.2\linewidth]{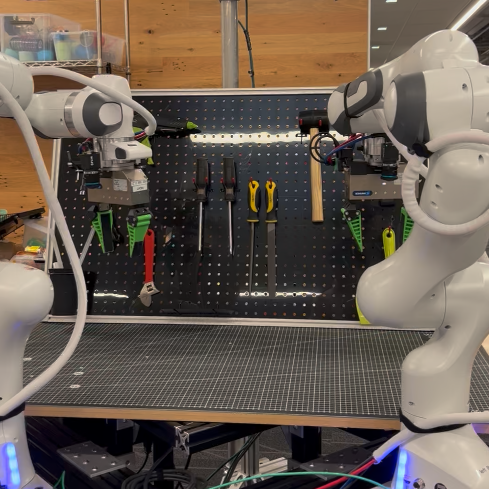} \\

    \small\textit{\shortstack{a red and black robot\\arm with a red handle}} &
  \small\textit{\shortstack{a cat sitting on the\\floor looking at the camera}} &
  \small\textit{\shortstack{a dog with a leash\\on a bench}} &
  \small\textit{\shortstack{a robot is working on\\a project in a workshop}} \\
  \end{tabular}
  \caption{VLM 1.1B caption-only model predictions (greedy decoding). The model uses normalized, 224$\times$224 input images to generate the captions. Images were sampled from the authors' phones (+ logo) to avoid any contamination.}
  \label{fig:vlm_captions}
  \end{figure}

\begin{table}[h]
    \caption{LLM evaluation results on multiple-choice reasoning benchmarks.
    HS = HellaSwag, WG = WinoGrande, OBQA = OpenBookQA.
    See descriptions, references, and terms of use in Appendix~\ref{app:llm_bench}.}
    \label{tab:llm_benchmarks}
    \centering
    \begin{tabular}{lcccccccc}
    \textbf{Model} & \textbf{HS} & \textbf{MMLU} & \textbf{ARC-e} &
    \textbf{ARC-c} & \textbf{PIQA} & \textbf{WG} & \textbf{OBQA} &
    \textbf{BoolQ} \\
    \midrule
    \foundryLLM{} (800B tokens) & 64.3 & 26.0 & 70.3 & 37.0 & 75.8 & 60.9 & 40.0 & 63.2 \\
    \foundryLLM{} (1T tokens)   & 66.7 & 26.6 & 71.7 & 39.3 & 77.5 & 62.6 & 40.8 & 65.4 \\
    \bottomrule
    \end{tabular}
\end{table}
\newpage
\paragraph{\textbf{VLM training}} We add a 86 million parameter randomly initialized vision transformer (ViT)~\cite{dosovitskiy2021imageworth16x16words}, with a similar architecture as CLIP~\cite{radford2021learning}, to encode ($224\times224$) input images. A pixel-shuffle~\cite{marafioti2025smolvlm}, operation is used as pooling to reduce the sequence length of the image.
We assemble the ViT and pooling with the previously pre-trained 1.2B LLM at 800B tokens of training -- before the learning rate cooldown, following recommendations from~\cite{keh2025should}. The VLM is trained with 200M samples of the openly available DataCompDR-1B ~\cite{gadre2023datacomp}\footnote{Image links from this dataset are known to break, which limits exact reproducibility.}. Our results are reported in Table~\ref{tab:vlm_coco_full} as evidence of end-to-end training functionality rather than as a claim of optimal performance. We also include qualitative examples in Figure~\ref{fig:vlm_captions}.

Although in this instance we use a randomly initialized ViT and the in-house LLM, both could instead be replaced by off-the-shelf pre-trained components such as SigLIP~\cite{zhai2023sigmoid} or DINO~\cite{oquab2023dinov2, simeoni2025cijo} which would likely lead to improved model performance. Alternatively, the VLM itself can take advantage of pre-trained backbones such as PaliGemma2~\cite{beyer_paligemma_2024} or Qwen3-VL~\cite{Qwen-VL}; this is precisely the route we take for \foundryQwenVLA{} in Section~\ref{sec:vlaf-qwen3}. Here we show that VLA Foundry supports all stages of training and can produce a functional VLM backbone, giving full control to users to experiment with known training data and procedures, modify architectures, and train or fine-tune any part of the model.

\begin{table}[h]
  \centering
  \caption{COCO\_VAL captioning evaluation. BLEU-n: Measures n-gram overlap between the generated caption and the references, ROUGE-L: Measures longest common subsequences, CIDEr: Measures weighted n-gram similarity (with n=1-4) so distinctive, informative phrases count more than common ones. }
  \label{tab:vlm_coco_full}
  \begin{tabular}{lcccccc}
      \textbf{Model} & \textbf{BLEU-1} & \textbf{BLEU-2} & \textbf{BLEU-3} & \textbf{BLEU-4}
      & \textbf{ROUGE\_L} & \textbf{CIDEr} \\
      \midrule
      \foundryVLM{} 165M & 57.25 & 37.12 & 23.23 & 14.44 & 37.13 & 50.17 \\
      \foundryVLM{} 200M & 58.64 & 38.62 & 24.49 & 15.57 & 38.17 & 55.14 \\
      \bottomrule
  \end{tabular}
\end{table}

\begin{figure}[H]
    \centering
    \includegraphics[width=1\linewidth, trim={4cm 4cm 4cm 4cm}, clip]{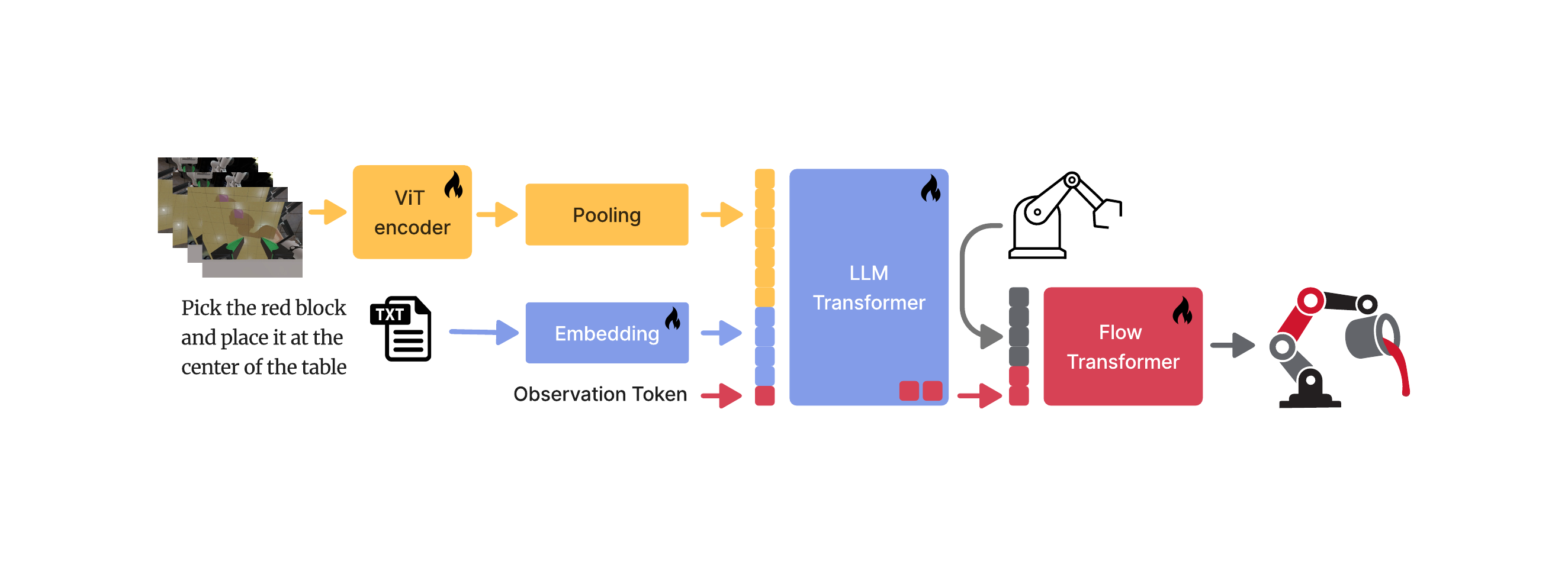}
    \caption{\foundryVLA{} architecture. Four images over two timesteps each are fed into the same ViT image encoder. For each of the 8 images, the result is pooled with ``pixel-shuffle''~\cite{marafioti2025smolvlm} (see appendix~\ref{app:pixel_shuffle}) and projected into the embedding space of the LLM. An additional observation token is appended to the sequence. The LLM embedding of the last layer matching this token is passed to a flow transformer with a noised action sequence. The flow transformer outputs the predicted denoising direction.}
    \label{fig:vla_archi}
\end{figure}

\paragraph{\textbf{VLA training}}
\label{sec:vla-data-and-arch}
We define the VLA architecture on top of the previous VLM (Figure~\ref{fig:vla_archi}). To extend the VLM architecture to to predict robot actions, we begin by adding a new \textit{observation} token to the LLM vocabulary. The VLM input sequence is composed of images and a text describing a task as well as the new observation token, in that order. The embedded sequence that is fed to the LLM part of the VLM is composed of the concatenated embedded image patches from multiple images, embedded text tokens, and the embedded observation token.
The hidden features of the last $N$ (in our experiments, 4) layers of the VLM matching the observation token are used to condition a flow transformer that denoises an action sequence. This action head is a 325 million parameter transformer with the same architecture as the LLM (except a vocabulary size of 0). Its input sequence is composed of the concatenated hidden features from the VLM, optionally, the proprioception encoded with a linear layer and the noised action sequence also encoded by a linear layer, in that order. The output action sequence is trained with the flow-matching objective~\cite{lipman2022flow}. We denote this model \foundryVLA{}.

We train \foundryVLA{} models on a data mixtures consisting of both simulated and real teleoperated demonstrations from stationary bimanual manipulation stations described in our previous work \lbm{}~\cite{lbmtri2025}. The data mix features 42 tasks in simulation and 361 tasks in the real world; 39 tasks are replicated in both real and simulation with copies of the stations and manipulands. Unlike our previous work we do not train on open-sourced data such as OXE~\cite{open_x_embodiment_rt_x_2023} or data collected with a universal manipulation device (UMI)~\cite{chi2024universal}. Further details regarding the dataset, including number of episodes per benchmark task and differences from the dataset of \lbm{}, can be found in Section~\ref{sec:vla_dataset}. Unless otherwise noted, \foundryVLA{} and \foundryQwenVLA{} are trained on a multi-task mixture of both real and simulation data\footnote{Download instructions for the processed LBM simulation data can be found in the codebase.}. We additionally train multi-task variants of \foundryVLA{} on simulation-only and real-only subsets, yielding \foundryVLASim{} and \foundryVLAReal{} respectively; these are used for the ablations in Section~\ref{sec:sim-eval}.

\subsection{\foundryQwenVLA{}: Leveraging a Strong VLM Backbone}
\label{sec:vlaf-qwen3}

A key design principle of VLA Foundry is that architectural components can be swapped with minimal effort. To exercise this, we also train a VLA with the pretrained Qwen3-VL 2B model~\cite{bai2025qwen3} as backbone. We reuse the same architecture as \foundryVLA{} for the action flow transformer and train on the full real and simulated data mixture. We denote this model \foundryQwenVLA{}.

\begin{figure}[htbp]
    \centering
    \includegraphics[width=\textwidth]{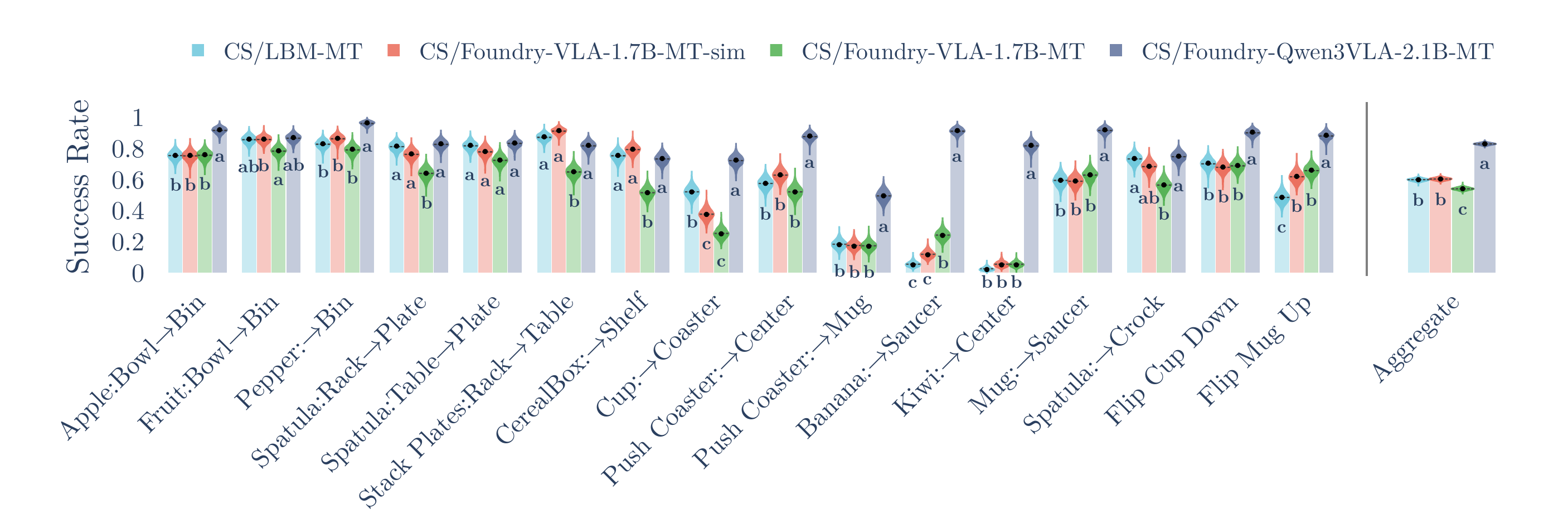}
    \caption{We compare our multi-task models—\foundryVLASim{}, \foundryVLA-\textsc{full}{}, and \foundryQwenVLA{}—against the \lbmmt{}~\cite{lbmtri2025} multi-task model on a set of seen tasks in \lbmevalcs{}. In aggregate, \lbmmt{} and \foundryVLASim{} are on par, while \foundryQwenVLA{} far outperforms the rest. Note that here only \foundryVLA-\textsc{full}{} and \foundryQwenVLA{} share the same exact robot training data; for more details refer to Section \ref{sec:lbm-vs-foundry}.}
    \label{fig:lbm_vs_foundry}
\end{figure}

The performance of \foundryQwenVLA{} demonstrates that a stronger and larger VLM backbone yields stronger VLA performance: \foundryQwenVLA{} improves over \foundryVLA{} on the shared simulation benchmark and outperforms our prior closed-source multi-task \lbm{} policy in a statistically significant manner by more than 20 percentage points (Figure~\ref{fig:lbm_vs_foundry}). Moreover, we show that the traditional VLM$\rightarrow$VLA recipe can be reproduced efficiently inside VLA Foundry, on the same training loop, dataloader, and preprocessing pipeline used for the from-scratch run, so practitioners do not need a separate training stack to adopt off-the-shelf backbones.

\subsection{Simulation Evaluation Results}
\label{sec:sim-eval}

\begin{figure}[htbp]
    \centering
    \includegraphics[width=\textwidth]{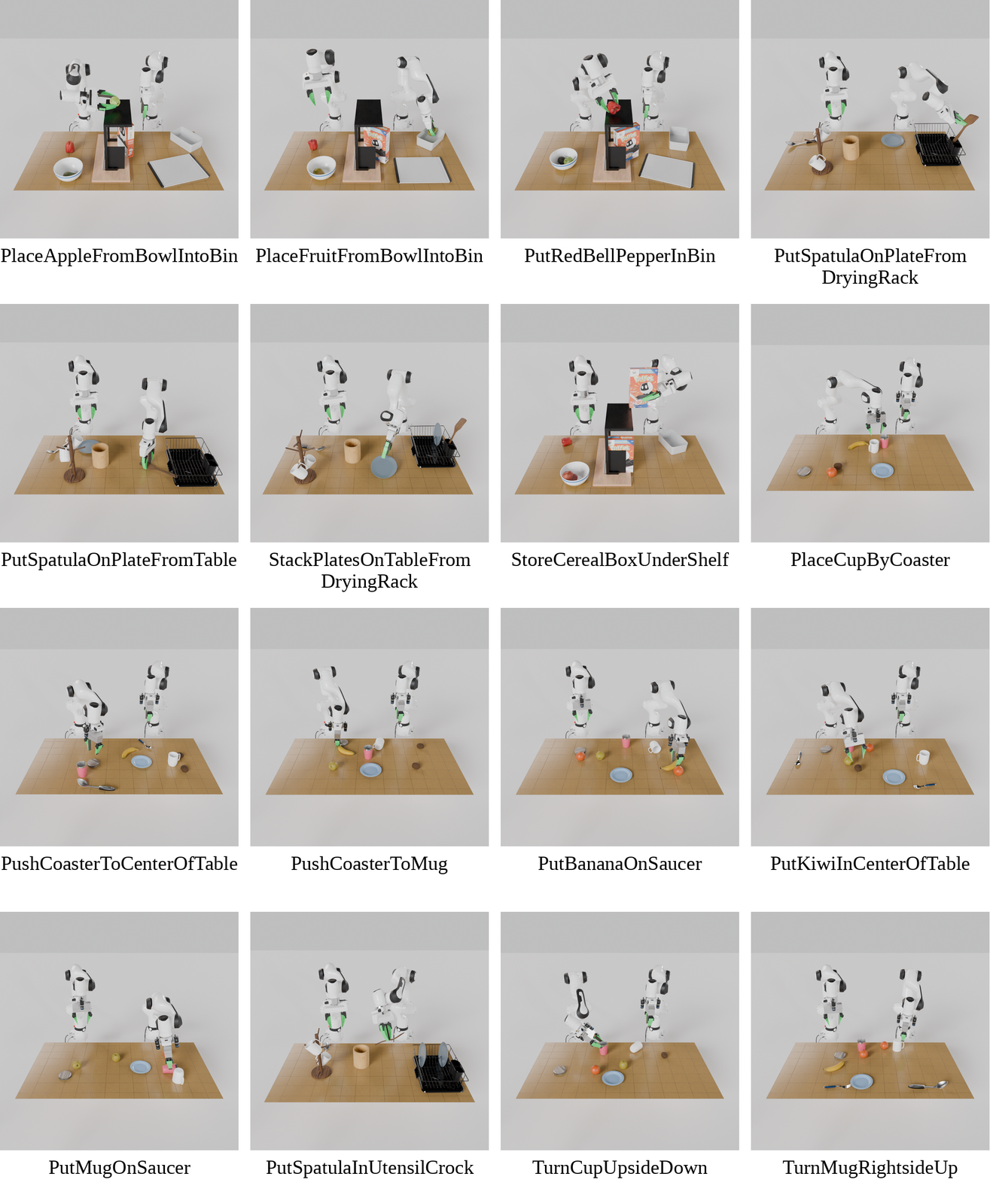}
    \caption{Overview of seen simulation evaluation tasks. The \lbmevaloss{} task suite spans tasks that require different qualities of manipulation capabilities: from pick-and-place to non-prehensile manipulation to bimanual coordination. Here, we show a single still from about the midpoint of a successful rollout from \foundryQwenVLA{}. Video versions of these images can be found at \foundryWebsite. We note that the images here build on top of~\cite{pfaff2025drakeblendertools}, where we re-light and re-render the Meshcat scenes at the desired frame rate from rollouts using Blender's Cycles after filtering out station geometry such as the external camera mounts, and table base for visual clarity; a representative example of sensor measurements actually used for model inference can be seen in Figure~\ref{fig:raw_sensor_measurements}. For a comparable figure of failed rollouts, refer to Figure~\ref{fig:simulation_task_overview_failures}.}
    \label{fig:simulation_task_overview_successes}
\end{figure}

In line with LBM~\cite{lbmtri2025}, we evaluate our models on a set of 16 simulation tasks (see Figure~\ref{fig:simulation_task_overview_successes}) seen at training time, as well as 3 simulation tasks held out from training\footnote{We do not evaluate on the distribution shift variant of the benchmark or additional long horizon simulation tasks; we leave this for future work.}, and compare performance with the statistical analysis tools introduced in Section~\ref{sec:dashboard}. The tasks in the benchmark vary in complexity and manipulation modes: PutKiwiInCenterOfTable is a simple pick-and-place task, PutRedBellPepperInBin requires one arm to place the bell pepper onto the shelf and the other arm to retrieve the item and place it in the bin, TurnCupUpsideDown requires only one arm but uses a wider range of motion especially in end effector rotations, and PushCoasterToMug requires non-prehensile manipulation. We evaluate on both the closed-source benchmark \lbmevalcs{} from which results were reported in~\cite{lbmtri2025} and the later open-sourced version \lbmevaloss{}~\cite{lbm_eval2025}. Due to updates between the two versions, policy performance can vary substantially, as \lbmevaloss{} can be considered a distribution-shifted version of the former; a comparison between model performance on the closed-source evaluation and the open-sourced evaluation on selected checkpoints is provided in Figure \ref{fig:cs_vs_os_sim_results}. For brevity, we use the following notation:

\begin{itemize}
  \item \textbf{CS}: closed-source; the simulation environment used in~\cite{lbmtri2025}; 
  it is largely the same used for data collection
  \item \textbf{OSS}: open-source software; the simulation environment that is openly accessible from~\cite{lbm_eval2025}
  \item \textbf{ST}: single-task; the model is trained and evaluated on the same task
  \item \textbf{MT}: multi-task; the model is trained on multiple tasks (can be simulation, real, or both)
  \item \textbf{FT}: multi-task finetuned: a multi-task checkpoint that is finetuned on a specific evaluation task
\end{itemize}

For both ST and FT, each task is evaluated with a specific set of model weights while MT models are evaluated on all the tasks with the same weights. All experiments are done with an evaluation budget of 200 rollout episodes\footnote{The results in this report are collated from an initial run and a followup run to patch missing trials; the evaluation results were then combined by keeping the most recent simulated episode. Therefore, exactly 200 rollout episodes were collected for each model.}. Note that some simulation seeds can also result in immediate, default successes; the raw data to produce the violin plots is included in the codebase.

In the following sections, we first compare models trained in VLA Foundry to \lbm{} on the closed source simulator. We then compare ST, MT, and FT training results for \foundryVLA{} and \foundryQwenVLA{} on seen and unseen tasks. For details of the violin plots and the CLD letters, refer to Section~\ref{sec:dashboard}. Additional results can be found in Section \ref{sec:additional_sim_results}.

\subsubsection{Comparison with \lbm{}}
\label{sec:lbm-vs-foundry}
First, we compare our results with \lbm{}, a multi-task model from previous work \cite{lbmtri2025}. \lbm{} is a 566 million parameter model that is composed of a pre-trained CLIP model for text and image embedding and a diffusion transformer head; in contrast to \foundryQwenVLA{} and \foundryVLA{}, the \lbm{} action head architecture utilizes cross-attention for the diffusion conditioning. Additionally, the \lbm{} model uses all camera images, zero padding when cameras are not present in certain data, whereas \foundryQwenVLA{} and \foundryVLA{} use only the two wrist camera and two external camera images shared between the different simulation stations.


Figure~\ref{fig:lbm_vs_foundry} compares \foundryQwenVLA{}, \foundryVLA{}, and \foundryVLASim{}{} multi-task against \lbm{} multi-task on \lbmevalcs{}. In aggregate, \foundryQwenVLA{} outperforms multi-task \lbm{} in terms of task success by a wide margin, while \lbm{} and \foundryVLASim{} are statistically on par with each other. \foundryVLA{}{} is the worst of the four models considered. Section \ref{sec:sim-real-foundry-vla} includes further evaluation and discussion on the effect of data recipes in the context of \foundryVLA{}.

\subsubsection{Training Stage Comparisons}

\begin{figure}[h]
    \centering
    \begin{subfigure}[b]{\textwidth}
        \centering
        \includegraphics[width=\textwidth]{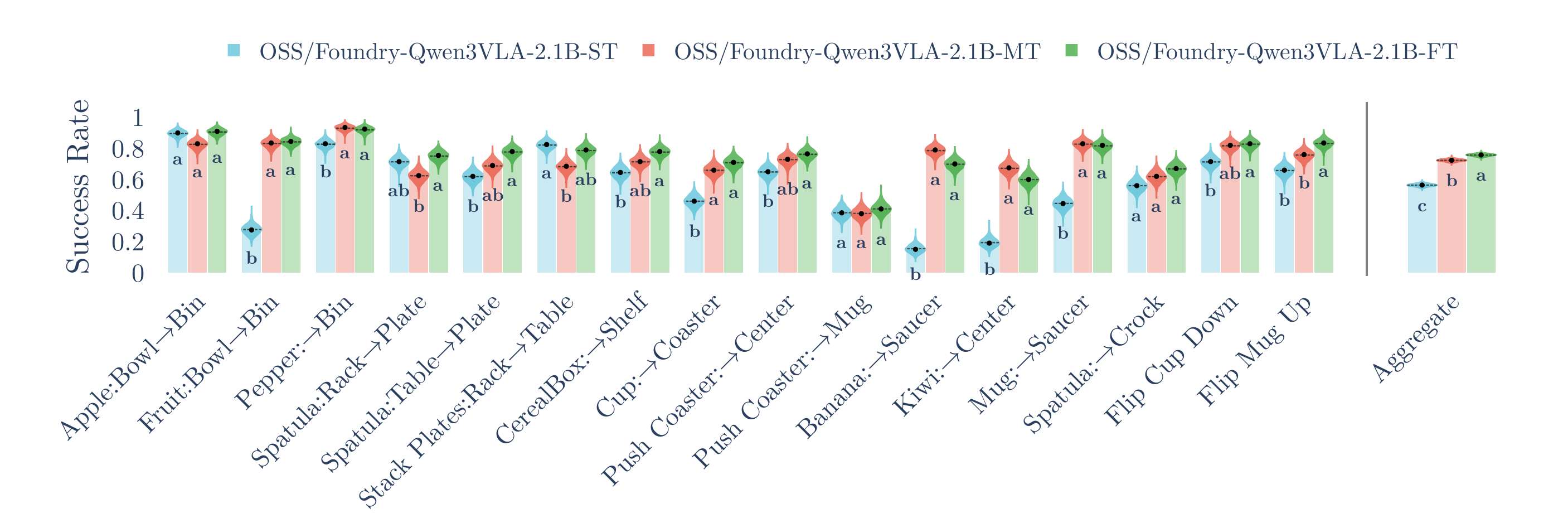}
        \caption{\foundryQwenVLA{} model series}
        \label{fig:foundry_qwen3vl2b}
    \end{subfigure}
    \vspace{0.5em}
    \begin{subfigure}[b]{\textwidth}
        \centering
        \includegraphics[width=\textwidth]{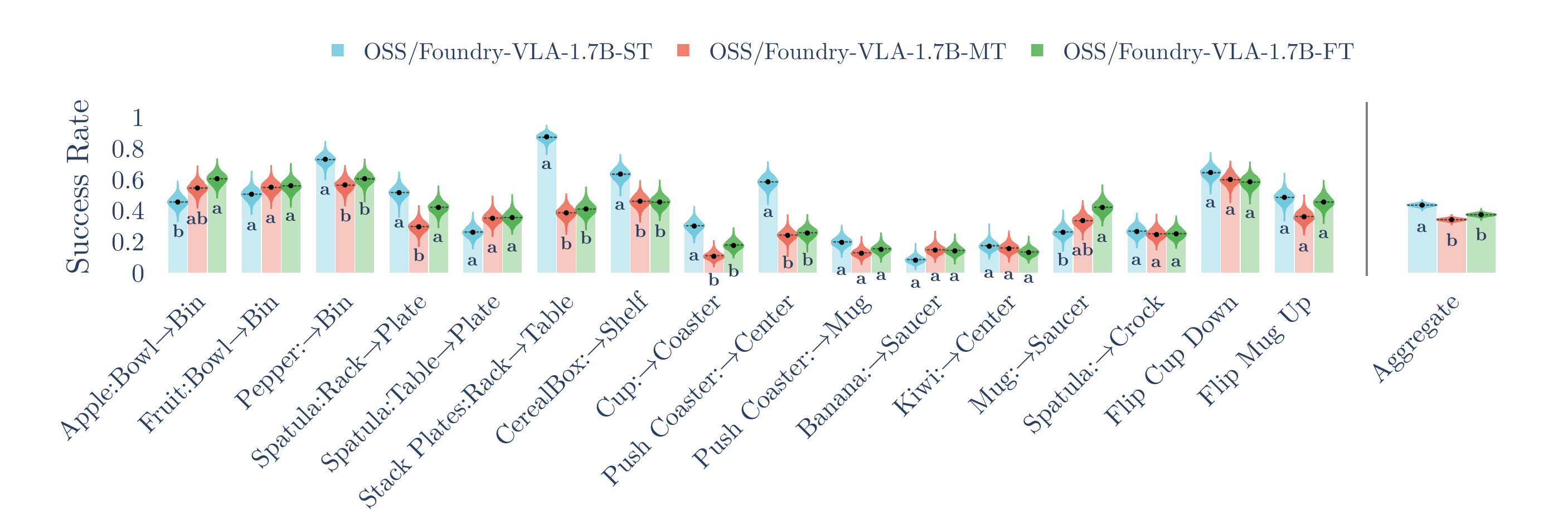}
        \caption{\foundryVLA{} model series}
        \label{fig:foundry_vla}
    \end{subfigure}
    \caption{Simulation results on \lbmevaloss{} (seen tasks). Aggregate performance increases from ST to MT to FT for the \foundryQwenVLA{} series; \foundryVLA{} performance is more mixed; the MT and FT variants are statistically worse than the ST.}
    \label{fig:foundryvla_qwenvla_seen}
\end{figure}

Figure~\ref{fig:foundryvla_qwenvla_seen}~(a) shows the results of \foundryQwenVLA{} at different training stages:
direct single-task training, multi-task training, and multi-task finetuned on each task.
After multi-task training on the simulation and real data, the \foundryQwenVLA{} model shows better performance than the single task training regime; finetuning the multi-task model on single seen tasks further improves performance in aggregate.\\
Figure~\ref{fig:foundryvla_qwenvla_seen}~(b) shows the same results from \foundryVLA{}. We see that while for some tasks such as \texttt{Apple:Bowl $\rightarrow$ Bin} the finetuned model outperforms the single task model, the opposite is true for other tasks such as \texttt{Stack Plates:Rack $\rightarrow$ Table}; in aggregate, the multi-task training and finetuning are statistically worse than the single task model.

\begin{figure}[h]
    \centering
    \begin{subfigure}[b]{0.48\textwidth}
        \centering
        \includegraphics[width=\textwidth]{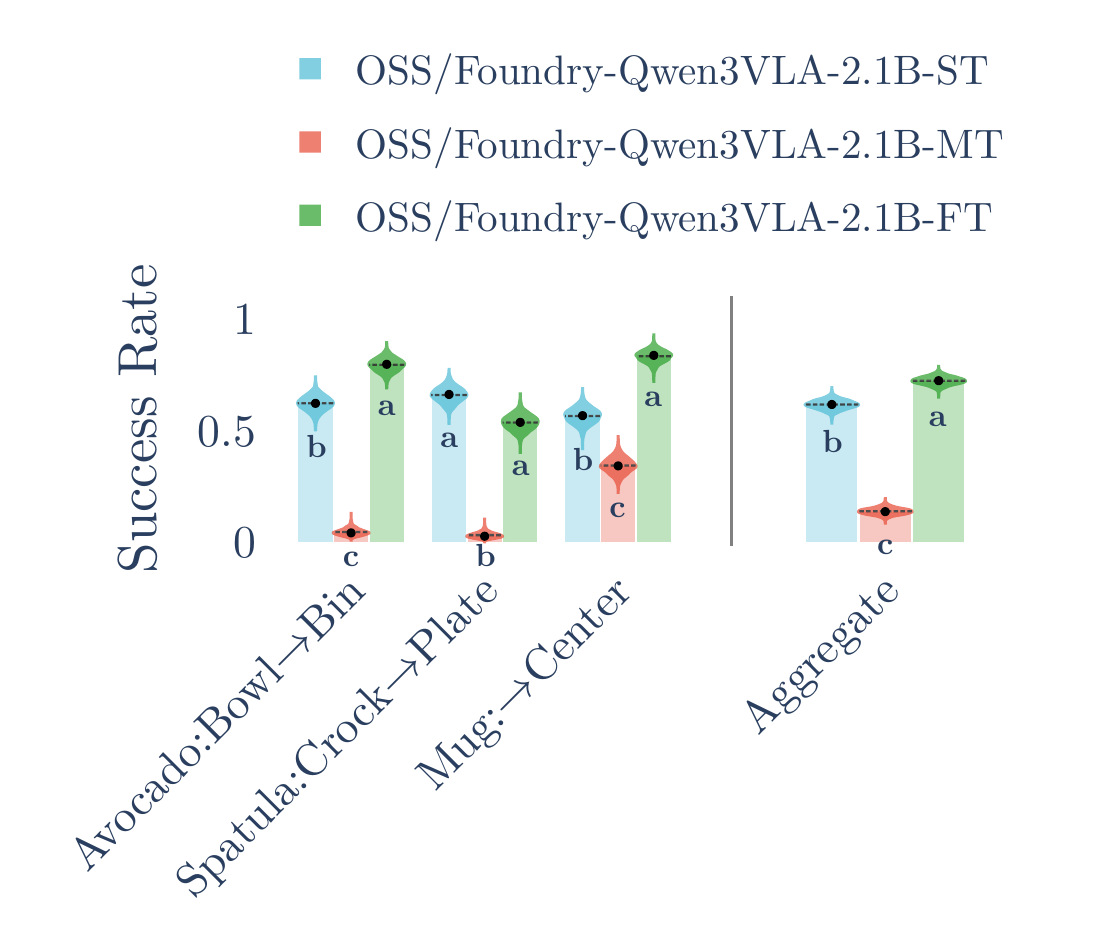}
        \caption{\foundryQwenVLA{} model series}
        \label{fig:foundry_qwen3vl2b_unseen}
    \end{subfigure}
    \hfill
    \begin{subfigure}[b]{0.48\textwidth}
        \centering
        \includegraphics[width=\textwidth]{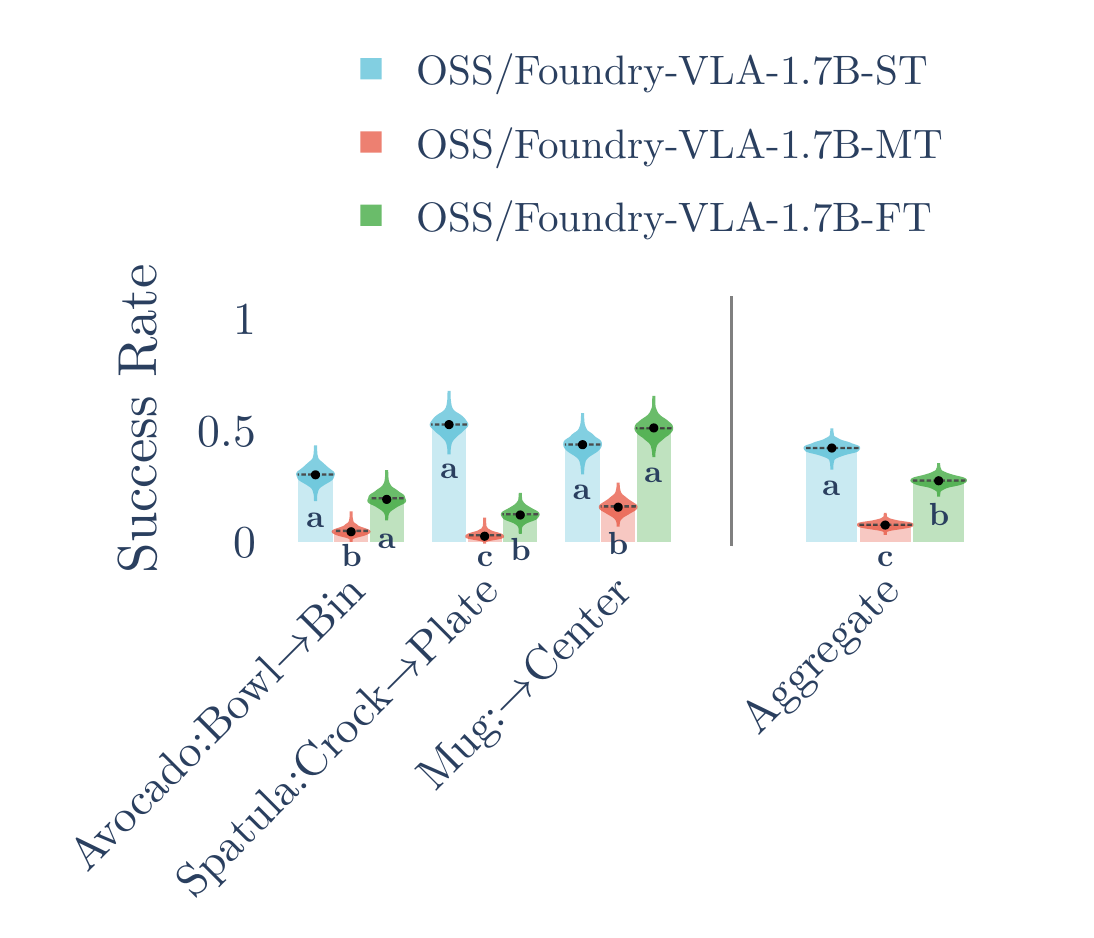}
        \caption{\foundryVLA{} model series}
        \label{fig:foundry_vla_unseen}
    \end{subfigure}
    \caption{Simulation results on \lbmevaloss{} (unseen tasks). Both \foundryVLA{} and \foundryQwenVLA{} demonstrate non-zero success rates 0-shot from real training to simulated evaluation.}
    \label{fig:foundryvla_qwenvla_unseen}
\end{figure}

Figure \ref{fig:foundryvla_qwenvla_unseen} shows the same models but evaluated on the 3 held-out tasks that are not part of the multi-task dataset. In both multi-task models, we observe some small amount of zero-shot generalization. However, while finetuning the multi-task \foundryQwenVLA{} model results in better performance than the single task variant in aggregate, the same is not true for \foundryVLA{}. These results are consistent with the hypothesis that stronger backbones can result in improved policy outcomes.

\subsubsection{Data Subset Comparisons}
\label{sec:sim-real-foundry-vla}
\begin{figure}
    \centering
    \includegraphics[width=1\linewidth]{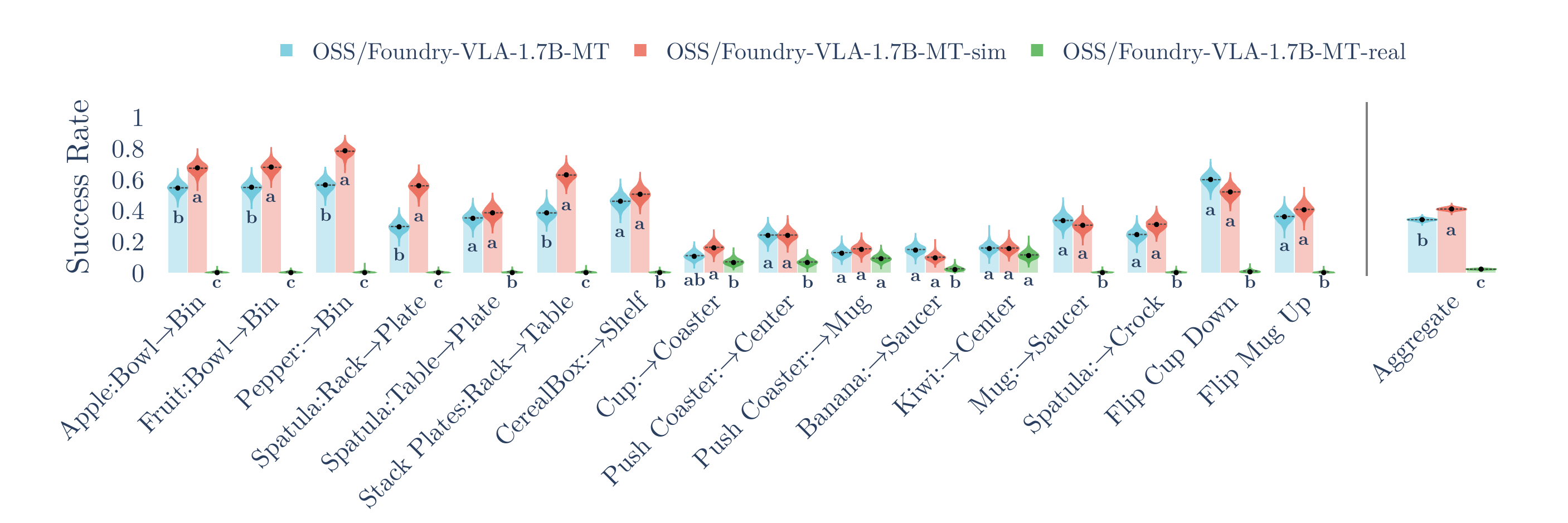}
    \caption{Simulation results of our three multitask \foundryVLA{} variants: trained on simulated data \foundryVLASim{}, real data \foundryVLAReal{}, and both combined \foundryVLAMT{}. }
    \label{fig:foundry-vla_data}
\end{figure}

To isolate the contribution of each data source, we additionally compare the results of training \foundryVLA{} on three subsets of data: simulation only, real robot only, and both combined.
 Simulation results of multi-task models trained on each of the 3 subsets are given in Figure~\ref{fig:foundry-vla_data}. All three models were trained for the same amount of compute but different amounts of data, i.e.,  the simulation only model was trained on more epochs of the same data. As expected the real-only training shows almost 0\% success rate because the simulation environment is out of its training distribution. Similar to Figure \ref{fig:lbm_vs_foundry}, the simulation only variant \foundryVLASim{}{} performs the best in aggregate. The number of episodes used to finetune the seen tasks can be found in Table~\ref{tab:data_per_task_breakdown}. Potential hypotheseses for the slightly worse performance compared to \foundryVLASim{} include model undertraining or the representational power of the model being split between real and simulated tasks; we leave further investigation to future work.

\section{Conclusions}
\label{sec:conclusion}

\paragraph{\textbf{Limitations}}
This initial release reflects deliberate choices in scope rather than framework constraints. Our reported evaluation is restricted to closed-loop LBM simulation on a narrow slice of embodiments, and we do not yet include real-hardware numbers; VLA Foundry's shared evaluation and dataloader abstractions are designed so that additional simulation suites (e.g., LIBERO, SimplerEnv, RoboCasa), new embodiments, and on-robot evaluation can be added without touching the core training stack. All experiments in this report use a flow-matching action head; while additional heads such as a diffusion policy are already implemented in the codebase, the action head is a modular component and integrating further variants -- for example, autoregressive discrete action tokenizations -- requires only a new head module rather than changes to the training loop or data pipeline. Finally, although VLA Foundry exposes the full LLM--VLM--VLA pipeline with probabilistic multi-modal mixing, we do not yet characterize optimal data recipes across stages, nor do we address safety, alignment, or failure-mode detection for embodied agents. We view these as open research directions that VLA Foundry is well-positioned to enable, and we invite the community to build on it to explore them.

\paragraph{\textbf{Conclusion}} 
In this technical report, we introduced \textbf{VLA Foundry}, an open-source framework that unifies LLM, VLM, and VLA training within a single codebase. The framework provides end-to-end control over the embodied-model pipeline -- from language pretraining through action learning -- with shared abstractions for data, configuration, training, and evaluation. Alongside the framework, we released two model types: \foundryVLA{}, trained fully from scratch through the LLM$\rightarrow$VLM$\rightarrow$VLA pipeline, and \foundryQwenVLA{}, built on a pretrained Qwen3-VL backbone with the same action head and training recipe.
On closed-loop LBM evaluation, \foundryVLASim{} is statistically on par with our prior closed-source LBM performance over our 16-tasks benchmark. \foundryQwenVLA{} outperforms both models with a wide margin of 23 percentage points on average. We demonstrated that VLA Foundry can be used to build VLAs both from-scratch and starting with a pretrained-backbone model. Together with the released checkpoints, the statistical comparison dashboard, and integration of \lbmevaloss{}, VLA Foundry's unified LLM--VLM--VLA stack enables the community to explore the design space that connects these stages -- training recipes, multi-modal data mixing, fusion architectures -- within a single codebase. We hope these tools will serve the community and that the community will contribute to their improvements.

\newpage
\subsection{Acknowledgements}
VLA Foundry would not be possible without the support of multiple teams and individuals at TRI.

Max Bajracharya managed the VLA team and provided guidance throughout the project. 

Mark Zolotas and Tim Chu provided feedback in various stages of the project and contributed quality-of-life improvements to the general infrastructure. We also thank Aykut Onol, Mengchao Zhang, Mark Zolotas, Naveen Kuppuswamy, and Sunny Sun for testing early versions of VLA Foundry on new embodiments. Ian McMahon and Jeremy Nimmer provided support for simulation evaluation. Andrew Beaulieu provided feedback to VLA Foundry and helped coordinate efforts with the TRI team in Cambridge.

Rishi Shah implemented small bugfixes and quality-of-life improvements, and helped test out VLA Foundry on various sim and hardware environments.

Richard Cheng, Chen Zou, Shanmuga Harikumar, Daiki Mori, Yukinori Kurahashi, and Takahiro Yamazaki provided support for testing VLA foundry on new simulation and mobile hardware environments. 

Chen Xu and Swati Gupta helped in early versions of our diffusion implementation. Pooja Kabra, Nagarjun Vinukonda, and David Berkowitz provided additional engineering support. We also thank Rhythm Syed, Jose Barreiros, Krishnan Srinivasan, and Blake Wulfe for support in various stages of the project. 

Satya Kotari provided compute infrastructure and AWS support. 

Nicholas Pfaff provided advice and code for rendering simulation rollouts.

Finally, we thank our robot teachers – Emma Dixon, Christopher Rodriguez, Derrick Seale, and Rudy Bravo for helping validate VLA Foundry on hardware. We also thank Patrick Miller and Masha Itkina for coordinating our data collection efforts. 

\subsection{Disclaimers}

Parts of the initial draft of the repo were taken from OpenLM \cite{open_lm}. Parts of the ViT implementation were taken from nanoVLM \cite{wiedmann2024nanovlm}.\\
The VLA Foundry codebase contains some code generated by LLMs. 


\printbibliography

@article{lbmtri2025,
  title={A Careful Examination of Large Behavior Models for Multitask Dexterous Manipulation}, 
  author={TRI LBM Team and Jose Barreiros and Andrew Beaulieu and Aditya Bhat and Rick Cory and Eric Cousineau and Hongkai Dai and Ching-Hsin Fang and Kunimatsu Hashimoto and Muhammad Zubair Irshad and Masha Itkina and Naveen Kuppuswamy and Kuan-Hui Lee and Katherine Liu and Dale McConachie and Ian McMahon and Haruki Nishimura and Calder Phillips-Grafflin and Charles Richter and Paarth Shah and Krishnan Srinivasan and Blake Wulfe and Chen Xu and Mengchao Zhang and Alex Alspach and Maya Angeles and Kushal Arora and Vitor Campagnolo Guizilini and Alejandro Castro and Dian Chen and Ting-Sheng Chu and Sam Creasey and Sean Curtis and Richard Denitto and Emma Dixon and Eric Dusel and Matthew Ferreira and Aimee Goncalves and Grant Gould and Damrong Guoy and Swati Gupta and Xuchen Han and Kyle Hatch and Brendan Hathaway and Allison Henry and Hillel Hochsztein and Phoebe Horgan and Shun Iwase and Donovon Jackson and Siddharth Karamcheti and Sedrick Keh and Joseph Masterjohn and Jean Mercat and Patrick Miller and Paul Mitiguy and Tony Nguyen and Jeremy Nimmer and Yuki Noguchi and Reko Ong and Aykut Onol and Owen Pfannenstiehl and Richard Poyner and Leticia Priebe Mendes Rocha and Gordon Richardson and Christopher Rodriguez and Derick Seale and Michael Sherman and Mariah Smith-Jones and David Tago and Pavel Tokmakov and Matthew Tran and Basile Van Hoorick and Igor Vasiljevic and Sergey Zakharov and Mark Zolotas and Rares Ambrus and Kerri Fetzer-Borelli and Benjamin Burchfiel and Hadas Kress-Gazit and Siyuan Feng and Stacie Ford and Russ Tedrake},
  year={2025},
  eprint={2507.05331},
  archivePrefix={arXiv},
  primaryClass={cs.RO},
  url={https://arxiv.org/abs/2507.05331}, 
}

@misc{open_lm,
  author = {Gururangan, Suchin and Wortsman, Mitchell and Gadre, Samir Yitzhak and Dave, Achal and Kilian, Maciej and Shi, Weijia and Mercat, Jean and Smyrnis, Georgios and Ilharco, Gabriel and Jordan, Matt and Heckel, Reinhard and Dimakis, Alex and Farhadi, Ali and Shankar, Vaishaal and Schmidt, Ludwig},
  title = {{open\_lm}:  a minimal but performative language modeling (LM) repository},
  year = {2023},
  note = {GitHub repository},
  url = {https://github.com/mlfoundations/open_lm/}
}

@misc{wiedmann2024nanovlm,
  author = {Luis Wiedmann and Juyoung Suk},
  title = {nanoVLM: The simplest repository to train your VLM in pure PyTorch},
  year = {2024},
  publisher = {GitHub},
  journal = {GitHub repository},
  howpublished = {\url{https://github.com/huggingface/nanoVLM}}
}

@inproceedings{karamcheti2024prismatic,
  title = {Prismatic VLMs: Investigating the Design Space of Visually-Conditioned Language Models},
  author = {Siddharth Karamcheti and Suraj Nair and Ashwin Balakrishna and Percy Liang and Thomas Kollar and Dorsa Sadigh},
  booktitle = {International Conference on Machine Learning (ICML)},
  year = {2024},
}

@article{liu2023visual,
  title={Visual instruction tuning},
  author={Liu, Haotian and Li, Chunyuan and Wu, Qingyang and Lee, Yong Jae},
  journal={Advances in neural information processing systems},
  volume={36},
  pages={34892--34916},
  year={2023}
}

@article{gadre2023datacomp,
  title={Datacomp: In search of the next generation of multimodal datasets},
  author={Gadre, Samir Yitzhak and Ilharco, Gabriel and Fang, Alex and Hayase, Jonathan and Smyrnis, Georgios and Nguyen, Thao and Marten, Ryan and Wortsman, Mitchell and Ghosh, Dhruba and Zhang, Jieyu and others},
  journal={Advances in Neural Information Processing Systems},
  volume={36},
  pages={27092--27112},
  year={2023}
}

@article{kim2024openvla,
  title={Openvla: An open-source vision-language-action model},
  author={Kim, Moo Jin and Pertsch, Karl and Karamcheti, Siddharth and Xiao, Ted and Balakrishna, Ashwin and Nair, Suraj and Rafailov, Rafael and Foster, Ethan and Lam, Grace and Sanketi, Pannag and others},
  journal={arXiv preprint arXiv:2406.09246},
  year={2024}
}

@article{keh2025should,
  title={Should VLMs be Pre-trained with Image Data?},
  author={Keh, Sedrick and Mercat, Jean and Gadre, Samir Yitzhak and Arora, Kushal and Vasiljevic, Igor and Burchfiel, Benjamin and Song, Shuran and Tedrake, Russ and Kollar, Thomas and Schmidt, Ludwig and others},
  journal={arXiv preprint arXiv:2503.07603},
  year={2025}
}

@misc{cadene2024lerobot,
    author = {Cadene, Remi and Alibert, Simon and Soare, Alexander and Gallouedec, Quentin and Zouitine, Adil and Palma, Steven and Kooijmans, Pepijn and Aractingi, Michel and Shukor, Mustafa and Aubakirova, Dana and Russi, Martino and Capuano, Francesco and Pascal, Caroline and Choghari, Jade and Moss, Jess and Wolf, Thomas},
    title = {LeRobot: State-of-the-art Machine Learning for Real-World Robotics in Pytorch},
    howpublished = "\url{https://github.com/huggingface/lerobot}",
    year = {2024}
}

@misc{draccus,
  author       = {{marin-community}},
  title        = {Draccus: Configuration with Dataclasses+YAML+Argparse},
  year         = {2026},
  howpublished = {\url{https://github.com/marin-community/draccus}},
}

@inproceedings{chen2024internvl,
  title={Internvl: Scaling up vision foundation models and aligning for generic visual-linguistic tasks},
  author={Chen, Zhe and Wu, Jiannan and Wang, Wenhai and Su, Weijie and Chen, Guo and Xing, Sen and Zhong, Muyan and Zhang, Qinglong and Zhu, Xizhou and Lu, Lewei and others},
  booktitle={Proceedings of the IEEE/CVF Conference on Computer Vision and Pattern Recognition},
  pages={24185--24198},
  year={2024}
}

@article{awadalla2023openflamingo,
  title={OpenFlamingo: An Open-Source Framework for Training Large Autoregressive Vision-Language Models},
  author={Anas Awadalla and Irena Gao and Josh Gardner and Jack Hessel and Yusuf Hanafy and Wanrong Zhu and Kalyani Marathe and Yonatan Bitton and Samir Gadre and Shiori Sagawa and Jenia Jitsev and Simon Kornblith and Pang Wei Koh and Gabriel Ilharco and Mitchell Wortsman and Ludwig Schmidt},
  journal={arXiv preprint arXiv:2308.01390},
  year={2023}
}

@article{Qwen-VL,
  title={Qwen-VL: A Versatile Vision-Language Model for Understanding, Localization, Text Reading, and Beyond},
  author={Bai, Jinze and Bai, Shuai and Yang, Shusheng and Wang, Shijie and Tan, Sinan and Wang, Peng and Lin, Junyang and Zhou, Chang and Zhou, Jingren},
  journal={arXiv preprint arXiv:2308.12966},
  year={2023}
}

@misc{black2025pi05,
  title         = {{$\pi_{0.5}$}: a Vision-Language-Action Model with Open-World Generalization},
  author        = {{Physical Intelligence} and Kevin Black and Noah Brown and James Darpinian and
                   Karan Dhabalia and Danny Driess and Adnan Esmail and Michael Equi and
                   Chelsea Finn and Niccolo Fusai and Manuel Y. Galliker and Dibya Ghosh and
                   Lachy Groom and Karol Hausman and Brian Ichter and Szymon Jakubczak and
                   Tim Jones and Liyiming Ke and Devin LeBlanc and Sergey Levine and
                   Adrian Li-Bell and Mohith Mothukuri and Suraj Nair and Karl Pertsch and
                   Allen Z. Ren and Lucy Xiaoyang Shi and Laura Smith and
                   Jost Tobias Springenberg and Kyle Stachowicz and James Tanner and
                   Quan Vuong and Homer Walke and Anna Walling and Haohuan Wang and
                   Lili Yu and Ury Zhilinsky},
  year          = {2025},
  eprint        = {2504.16054},
  archivePrefix = {arXiv},
  primaryClass  = {cs.RO},
  url           = {https://arxiv.org/abs/2504.16054}
}

@misc{intelligence2025pi06star,
  title         = {{$\pi^{*}_{0.6}$}: a VLA That Learns From Experience},
  author        = {{Physical Intelligence} and Ali Amin and Raichelle Aniceto and
                   Ashwin Balakrishna and Kevin Black and Ken Conley and Grace Connors and
                   James Darpinian and Karan Dhabalia and Jared DiCarlo and Danny Driess and
                   Michael Equi and Adnan Esmail and Yunhao Fang and Chelsea Finn and
                   Catherine Glossop and Thomas Godden and Ivan Goryachev and Lachy Groom and
                   Hunter Hancock and Karol Hausman and Gashon Hussein and Brian Ichter and
                   Szymon Jakubczak and Rowan Jen and Tim Jones and Ben Katz and
                   Liyiming Ke and Chandra Kuchi and Marinda Lamb and Devin LeBlanc and
                   Sergey Levine and Adrian Li-Bell and Yao Lu and Vishnu Mano and
                   Mohith Mothukuri and Suraj Nair and Karl Pertsch and Allen Z. Ren and
                   Charvi Sharma and Lucy Xiaoyang Shi and Laura Smith and
                   Jost Tobias Springenberg and Kyle Stachowicz and Will Stoeckle and
                   Alex Swerdlow and James Tanner and Marcel Torne and Quan Vuong and
                   Anna Walling and Haohuan Wang and Blake Williams and Sukwon Yoo and
                   Lili Yu and Ury Zhilinsky and Zhiyuan Zhou},
  year          = {2025},
  eprint        = {2511.14759},
  archivePrefix = {arXiv},
  primaryClass  = {cs.LG},
  url           = {https://arxiv.org/abs/2511.14759}
}

@inproceedings{gr00tn1_2025,
  archivePrefix = {arxiv},
  eprint     = {2503.14734},
  title      = {{GR00T} {N1}: An Open Foundation Model for Generalist Humanoid Robots},
  author     = {NVIDIA and Johan Bjorck and Fernando Castañeda, Nikita Cherniadev and Xingye Da and Runyu Ding and Linxi "Jim" Fan and Yu Fang and Dieter Fox and Fengyuan Hu and Spencer Huang and Joel Jang and Zhenyu Jiang and Jan Kautz and Kaushil Kundalia and Lawrence Lao and Zhiqi Li and Zongyu Lin and Kevin Lin and Guilin Liu and Edith Llontop and Loic Magne and Ajay Mandlekar and Avnish Narayan and Soroush Nasiriany and Scott Reed and You Liang Tan and Guanzhi Wang and Zu Wang and Jing Wang and Qi Wang and Jiannan Xiang and Yuqi Xie and Yinzhen Xu and Zhenjia Xu and Seonghyeon Ye and Zhiding Yu and Ao Zhang and Hao Zhang and Yizhou Zhao and Ruijie Zheng and Yuke Zhu},
  month      = {March},
  year       = {2025},
  booktitle  = {ArXiv Preprint},
}

@misc{molmoact2025,
      title={MolmoAct: Action Reasoning Models that can Reason in Space}, 
      author={Jason Lee and Jiafei Duan and Haoquan Fang and Yuquan Deng and Shuo Liu and Boyang Li and Bohan Fang and Jieyu Zhang and Yi Ru Wang and Sangho Lee and Winson Han and Wilbert Pumacay and Angelica Wu and Rose Hendrix and Karen Farley and Eli VanderBilt and Ali Farhadi and Dieter Fox and Ranjay Krishna},
      year={2025},
      eprint={2508.07917},
      archivePrefix={arXiv},
      primaryClass={cs.RO},
      url={https://arxiv.org/abs/2508.07917}
}

@article{shukor2025smolvla,
  title   = {SmolVLA: A vision-language-action model for affordable and efficient robotics},
  author  = {Shukor, Mustafa and Aubakirova, Dana and Capuano, Francesco and Kooijmans, Pepijn and Palma, Steven and Zouitine, Adil and Aractingi, Michel and Pascal, Caroline and Russi, Martino and Marafioti, Andres and Alibert, Simon and Cord, Matthieu and Wolf, Thomas and Cadene, Remi},
  year    = {2025},
  journal = {arXiv preprint},
  eprint  = {2506.01844},
  archivePrefix = {arXiv},
  primaryClass  = {cs.RO}
}

@misc{dreamzero_ye2026worldactionmodelszeroshot,
      title={World Action Models are Zero-shot Policies}, 
      author={Seonghyeon Ye and Yunhao Ge and Kaiyuan Zheng and Shenyuan Gao and Sihyun Yu and George Kurian and Suneel Indupuru and You Liang Tan and Chuning Zhu and Jiannan Xiang and Ayaan Malik and Kyungmin Lee and William Liang and Nadun Ranawaka and Jiasheng Gu and Yinzhen Xu and Guanzhi Wang and Fengyuan Hu and Avnish Narayan and Johan Bjorck and Jing Wang and Gwanghyun Kim and Dantong Niu and Ruijie Zheng and Yuqi Xie and Jimmy Wu and Qi Wang and Ryan Julian and Danfei Xu and Yilun Du and Yevgen Chebotar and Scott Reed and Jan Kautz and Yuke Zhu and Linxi "Jim" Fan and Joel Jang},
      year={2026},
      eprint={2602.15922},
      archivePrefix={arXiv},
      primaryClass={cs.RO},
      url={https://arxiv.org/abs/2602.15922}, 
}

@article{lin2026holobrain,
  title={HoloBrain-0 Technical Report},
  author={Lin, Xuewu and Lin, Tianwei and Du, Yun and Xie, Hongyu and Jin, Yiwei and Li, Jiawei and Wu, Shijie and Wang, Qingze and Li, Mengdi  and Zhao, Mengao and Li, Ziang and Huang, Chaodong and Bi, Hongzhe and Huang, Lichao and Su, Zhizhong},
  journal={arXiv preprint arXiv:2602.12062},
  year={2026},
  url={https://arxiv.org/abs/2602.12062},
}

@misc{galahad2025vlascratch,
  title        = {{VLA-Scratch}: A Modular, Performant, Efficient Stack For Vision-Language-Action Models},
  author       = {EGalahad},
  year         = {2025},
  howpublished = {\url{https://github.com/EGalahad/vla-scratch}},
  note         = {GitHub repository}
}

@misc{physicalintelligence2025openpi,
  title        = {{openpi}: Open-Source Models and Packages for Robotics},
  author       = {{Physical Intelligence}},
  year         = {2025},
  howpublished = {\url{https://github.com/Physical-Intelligence/openpi}},
  note         = {GitHub repository. Apache-2.0 License}
}

@misc{aubakirova2025vlab,
  author = {Dana Aubakirova, Mustafa Shukor and Jade Cholgari and Leandro von Werra},
  title = {VLAb: Your Laboratory for Pretraining VLAs},
  year = {2025},
  publisher = {GitHub},
  journal = {GitHub repository},
  howpublished = {\url{https://github.com/huggingface/vlab}}
}

@misc{weng2026vlascratch,
  author       = {Weng, Haoyang and Lin, Haotian and Kalander, Marcus and Gu, Chenyang and Wang, Ziren and Yang, Rujia},
  title        = {VLA-Scratch: Modular, Performant and Efficient Stack},
  year         = {2026},
  howpublished = {\url{https://github.com/EGalahad/vla-scratch}},
  note         = {GitHub repository}
}

@article{dunning2019computing,
  title={Computing Extremely Accurate Quantiles Using t-Digests},
  author={Dunning, Ted and Ertl, Otmar},
  journal={arXiv preprint arXiv:1902.04023},
  year={2019},
  url={https://arxiv.org/abs/1902.04023}
}

@InProceedings{zhou2019continuity,
  author    = {Zhou, Yi and Barnes, Connelly and Lu, Jingwan and Yang, Jimei and Li, Hao},
  title     = {On the Continuity of Rotation Representations in Neural Networks},
  booktitle = {Proceedings of the IEEE/CVF Conference on Computer Vision and Pattern Recognition (CVPR)},
  month     = {June},
  year      = {2019},
  pages     = {5745--5753}
}

@InProceedings{Zhao-RSS-23,
  author    = {Tony Z. Zhao and Vikash Kumar and Sergey Levine and Chelsea Finn},
  title     = {{Learning Fine-Grained Bimanual Manipulation with Low-Cost Hardware}},
  booktitle = {Proceedings of Robotics: Science and Systems},
  year      = {2023},
  address   = {Daegu, Republic of Korea},
  month     = {July},
  doi       = {10.15607/RSS.2023.XIX.016}
}

@article{megatron-lm,
  title={Megatron-LM: Training Multi-Billion Parameter Language Models Using Model Parallelism},
  author={Shoeybi, Mohammad and Patwary, Mostofa and Puri, Raul and LeGresley, Patrick and Casper, Jared and Catanzaro, Bryan},
  journal={arXiv preprint arXiv:1909.08053},
  year={2019}
}

@misc{olmo20242olmo2furious,
      title={{2 OLMo 2 Furious}},
      author={{Team OLMo} and Pete Walsh and Luca Soldaini and Dirk Groeneveld and Kyle Lo and Shane Arora and Akshita Bhagia and Yuling Gu and Shengyi Huang and Matt Jordan and Nathan Lambert and Dustin Schwenk and Oyvind Tafjord and Taira Anderson and David Atkinson and Faeze Brahman and Christopher Clark and Pradeep Dasigi and Nouha Dziri and Michal Guerquin and Hamish Ivison and Pang Wei Koh and Jiacheng Liu and Saumya Malik and William Merrill and Lester James V. Miranda and Jacob Morrison and Tyler Murray and Crystal Nam and Valentina Pyatkin and Aman Rangapur and Michael Schmitz and Sam Skjonsberg and David Wadden and Christopher Wilhelm and Michael Wilson and Luke Zettlemoyer and Ali Farhadi and Noah A. Smith and Hannaneh Hajishirzi},
      year={2024},
      eprint={2501.00656},
      archivePrefix={arXiv},
      primaryClass={cs.CL},
      url={https://arxiv.org/abs/2501.00656},
}

@inproceedings{snyder2025step,
  title = {Is Your Imitation Learning Policy Better Than Mine? Policy Comparison with Near-Optimal Stopping},
  author = {Snyder, David and Hancock, Asher James and Badithela, Apurva and Dixon, Emma and Miller, Patrick and Ambrus, Rares Andrei and Majumdar, Anirudha and Itkina, Masha and Nishimura, Haruki},
  booktitle={Proceedings of the Robotics: Science and Systems Conference (RSS) XXI},
  year = {2025},
}

@article{piepho2004algorithm,
  title={An algorithm for a letter-based representation of all-pairwise comparisons},
  author={Piepho, Hans-Peter},
  journal={Journal of Computational and Graphical Statistics},
  volume={13},
  number={2},
  pages={456--466},
  year={2004},
  publisher={Taylor \& Francis}
}

@article{stefan2023big,
  title={Big little lies: A compendium and simulation of p-hacking strategies},
  author={Stefan, Angelika M and Sch{\"o}nbrodt, Felix D},
  journal={Royal Society Open Science},
  volume={10},
  number={2},
  year={2023},
  publisher={The Royal Society}
}

@article{barnard_significance_1947,
	title = {\href{https://doi.org/10.1093/biomet/34.1-2.123}{{Significance} {Tests} {for} 2×2 {Tables}}},
	volume = {34},
	issn = {0006-3444},
	doi = {10.1093/biomet/34.1-2.123},
	number = {1-2},
	urldate = {2025-01-20},
	journal = {Biometrika},
	author = {Barnard, G. A.},
	month = jan,
	year = {1947},
	pages = {123--138},
}

@misc{tri_2026_ab_testing,
  author       = {Nishimura, Haruki and Itkina, Masha},
  title        = {Statistical Thinking for Robot Policy Evaluation: From Rigorous A/B Testing to Effective Visualization},
  year         = {2026},
  url          = {https://medium.com/toyotaresearch/statistical-thinking-for-robot-policy-evaluation-from-rigorous-a-b-testing-to-effective-0ae886fbd68d},
  note         = {Accessed: 2026-04-17},
  howpublished = {Medium}
}

@misc{lerobot_unfolding_robotics,
  author = {Kooijmans, Pepijn and  Aractingi, Michel and Palma, Steven and Pascal, Caroline and Choghari, Jade and Meftah, Khalil and Russi, Martino and Rabault, Nicolas and Batto, Virgile and  Werra, Leandro von and Wolf, Thomas},
  title  = {Unfolding Robotics: The Open-Source Recipe for Teaching a Robot to Fold Your Clothes},
  year   = {2026},
  url    = {https://huggingface.co/spaces/lerobot/robot-folding},
  note   = {Accessed: 2026-04-17}
}

@misc{open_x_embodiment_rt_x_2023,
title={Open {X-E}mbodiment: Robotic Learning Datasets and {RT-X} Models},
author = {Open X-Embodiment Collaboration and Abby O'Neill and Abdul Rehman and Abhinav Gupta and Abhiram Maddukuri and Abhishek Gupta and Abhishek Padalkar and Abraham Lee and Acorn Pooley and Agrim Gupta and Ajay Mandlekar and Ajinkya Jain and Albert Tung and Alex Bewley and Alex Herzog and Alex Irpan and Alexander Khazatsky and Anant Rai and Anchit Gupta and Andrew Wang and Andrey Kolobov and Anikait Singh and Animesh Garg and Aniruddha Kembhavi and Annie Xie and Anthony Brohan and Antonin Raffin and Archit Sharma and Arefeh Yavary and Arhan Jain and Ashwin Balakrishna and Ayzaan Wahid and Ben Burgess-Limerick and Beomjoon Kim and Bernhard Schölkopf and Blake Wulfe and Brian Ichter and Cewu Lu and Charles Xu and Charlotte Le and Chelsea Finn and Chen Wang and Chenfeng Xu and Cheng Chi and Chenguang Huang and Christine Chan and Christopher Agia and Chuer Pan and Chuyuan Fu and Coline Devin and Danfei Xu and Daniel Morton and Danny Driess and Daphne Chen and Deepak Pathak and Dhruv Shah and Dieter Büchler and Dinesh Jayaraman and Dmitry Kalashnikov and Dorsa Sadigh and Edward Johns and Ethan Foster and Fangchen Liu and Federico Ceola and Fei Xia and Feiyu Zhao and Felipe Vieira Frujeri and Freek Stulp and Gaoyue Zhou and Gaurav S. Sukhatme and Gautam Salhotra and Ge Yan and Gilbert Feng and Giulio Schiavi and Glen Berseth and Gregory Kahn and Guangwen Yang and Guanzhi Wang and Hao Su and Hao-Shu Fang and Haochen Shi and Henghui Bao and Heni Ben Amor and Henrik I Christensen and Hiroki Furuta and Homanga Bharadhwaj and Homer Walke and Hongjie Fang and Huy Ha and Igor Mordatch and Ilija Radosavovic and Isabel Leal and Jacky Liang and Jad Abou-Chakra and Jaehyung Kim and Jaimyn Drake and Jan Peters and Jan Schneider and Jasmine Hsu and Jay Vakil and Jeannette Bohg and Jeffrey Bingham and Jeffrey Wu and Jensen Gao and Jiaheng Hu and Jiajun Wu and Jialin Wu and Jiankai Sun and Jianlan Luo and Jiayuan Gu and Jie Tan and Jihoon Oh and Jimmy Wu and Jingpei Lu and Jingyun Yang and Jitendra Malik and João Silvério and Joey Hejna and Jonathan Booher and Jonathan Tompson and Jonathan Yang and Jordi Salvador and Joseph J. Lim and Junhyek Han and Kaiyuan Wang and Kanishka Rao and Karl Pertsch and Karol Hausman and Keegan Go and Keerthana Gopalakrishnan and Ken Goldberg and Kendra Byrne and Kenneth Oslund and Kento Kawaharazuka and Kevin Black and Kevin Lin and Kevin Zhang and Kiana Ehsani and Kiran Lekkala and Kirsty Ellis and Krishan Rana and Krishnan Srinivasan and Kuan Fang and Kunal Pratap Singh and Kuo-Hao Zeng and Kyle Hatch and Kyle Hsu and Laurent Itti and Lawrence Yunliang Chen and Lerrel Pinto and Li Fei-Fei and Liam Tan and Linxi "Jim" Fan and Lionel Ott and Lisa Lee and Luca Weihs and Magnum Chen and Marion Lepert and Marius Memmel and Masayoshi Tomizuka and Masha Itkina and Mateo Guaman Castro and Max Spero and Maximilian Du and Michael Ahn and Michael C. Yip and Mingtong Zhang and Mingyu Ding and Minho Heo and Mohan Kumar Srirama and Mohit Sharma and Moo Jin Kim and Muhammad Zubair Irshad and Naoaki Kanazawa and Nicklas Hansen and Nicolas Heess and Nikhil J Joshi and Niko Suenderhauf and Ning Liu and Norman Di Palo and Nur Muhammad Mahi Shafiullah and Oier Mees and Oliver Kroemer and Osbert Bastani and Pannag R Sanketi and Patrick "Tree" Miller and Patrick Yin and Paul Wohlhart and Peng Xu and Peter David Fagan and Peter Mitrano and Pierre Sermanet and Pieter Abbeel and Priya Sundaresan and Qiuyu Chen and Quan Vuong and Rafael Rafailov and Ran Tian and Ria Doshi and Roberto Mart{'i}n-Mart{'i}n and Rohan Baijal and Rosario Scalise and Rose Hendrix and Roy Lin and Runjia Qian and Ruohan Zhang and Russell Mendonca and Rutav Shah and Ryan Hoque and Ryan Julian and Samuel Bustamante and Sean Kirmani and Sergey Levine and Shan Lin and Sherry Moore and Shikhar Bahl and Shivin Dass and Shubham Sonawani and Shubham Tulsiani and Shuran Song and Sichun Xu and Siddhant Haldar and Siddharth Karamcheti and Simeon Adebola and Simon Guist and Soroush Nasiriany and Stefan Schaal and Stefan Welker and Stephen Tian and Subramanian Ramamoorthy and Sudeep Dasari and Suneel Belkhale and Sungjae Park and Suraj Nair and Suvir Mirchandani and Takayuki Osa and Tanmay Gupta and Tatsuya Harada and Tatsuya Matsushima and Ted Xiao and Thomas Kollar and Tianhe Yu and Tianli Ding and Todor Davchev and Tony Z. Zhao and Travis Armstrong and Trevor Darrell and Trinity Chung and Vidhi Jain and Vikash Kumar and Vincent Vanhoucke and Vitor Guizilini and Wei Zhan and Wenxuan Zhou and Wolfram Burgard and Xi Chen and Xiangyu Chen and Xiaolong Wang and Xinghao Zhu and Xinyang Geng and Xiyuan Liu and Xu Liangwei and Xuanlin Li and Yansong Pang and Yao Lu and Yecheng Jason Ma and Yejin Kim and Yevgen Chebotar and Yifan Zhou and Yifeng Zhu and Yilin Wu and Ying Xu and Yixuan Wang and Yonatan Bisk and Yongqiang Dou and Yoonyoung Cho and Youngwoon Lee and Yuchen Cui and Yue Cao and Yueh-Hua Wu and Yujin Tang and Yuke Zhu and Yunchu Zhang and Yunfan Jiang and Yunshuang Li and Yunzhu Li and Yusuke Iwasawa and Yutaka Matsuo and Zehan Ma and Zhuo Xu and Zichen Jeff Cui and Zichen Zhang and Zipeng Fu and Zipeng Lin},
howpublished  = {\url{https://arxiv.org/abs/2310.08864}},
year = {2023},
}

@article{hu2024minicpm,
  title={Minicpm: Unveiling the potential of small language models with scalable training strategies},
  author={Hu, Shengding and Tu, Yuge and Han, Xu and He, Chaoqun and Cui, Ganqu and Long, Xiang and Zheng, Zhi and Fang, Yewei and Huang, Yuxiang and Zhao, Weilin and others},
  journal={arXiv preprint arXiv:2404.06395},
  year={2024}
}

@misc{dosovitskiy2021imageworth16x16words,
      title={An Image is Worth 16x16 Words: Transformers for Image Recognition at Scale}, 
      author={Alexey Dosovitskiy and Lucas Beyer and Alexander Kolesnikov and Dirk Weissenborn and Xiaohua Zhai and Thomas Unterthiner and Mostafa Dehghani and Matthias Minderer and Georg Heigold and Sylvain Gelly and Jakob Uszkoreit and Neil Houlsby},
      year={2021},
      eprint={2010.11929},
      archivePrefix={arXiv},
      primaryClass={cs.CV},
      url={https://arxiv.org/abs/2010.11929}, 
}

@article{lin2026systematic,
  title={A Systematic Study of Data Modalities and Strategies for Co-training Large Behavior Models for Robot Manipulation},
  author={Lin, Fanqi and Arora, Kushal and Mercat, Jean and Nishimura, Haruki and Shah, Paarth and Xu, Chen and Zhang, Mengchao and Zolotas, Mark and Angeles, Maya and Pfannenstiehl, Owen and others},
  journal={arXiv preprint arXiv:2602.01067},
  year={2026}
}

@inproceedings{zhai2023sigmoid,
  title={Sigmoid loss for language image pre-training},
  author={Zhai, Xiaohua and Mustafa, Basil and Kolesnikov, Alexander and Beyer, Lucas},
  booktitle={Proceedings of the IEEE/CVF international conference on computer vision},
  pages={11975--11986},
  year={2023}
}

@article{oquab2023dinov2,
  title={Dinov2: Learning robust visual features without supervision},
  author={Oquab, Maxime and Darcet, Timoth{\'e}e and Moutakanni, Th{\'e}o and Vo, Huy and Szafraniec, Marc and Khalidov, Vasil and Fernandez, Pierre and Haziza, Daniel and Massa, Francisco and El-Nouby, Alaaeldin and others},
  journal={arXiv preprint arXiv:2304.07193},
  year={2023}
}

@article{simeoni2025cijo,
  title={Dinov3},
  author={Sim{\'e}oni, Oriane and Vo, Huy V and Seitzer, Maximilian and Baldassarre, Federico and Oquab, Maxime},
  journal={arXiv preprint arXiv:2508.10104},
  year={2025}
}

@misc{beyer_paligemma_2024,
    title = {{PaliGemma}: {A} versatile {3B} {VLM} for transfer},
    shorttitle = {{PaliGemma}},
    url = {https://arxiv.org/abs/2407.07726v1},
    language = {en},
    urldate = {2024-09-05},
    journal = {arXiv.org},
    author = {Beyer, Lucas and Steiner, Andreas and Pinto, André Susano and Kolesnikov, Alexander and Wang, Xiao and Salz, Daniel and Neumann, Maxim and Alabdulmohsin, Ibrahim and Tschannen, Michael and Bugliarello, Emanuele and Unterthiner, Thomas and Keysers, Daniel and Koppula, Skanda and Liu, Fangyu and Grycner, Adam and Gritsenko, Alexey and Houlsby, Neil and Kumar, Manoj and Rong, Keran and Eisenschlos, Julian and Kabra, Rishabh and Bauer, Matthias and Bošnjak, Matko and Chen, Xi and Minderer, Matthias and Voigtlaender, Paul and Bica, Ioana and Balazevic, Ivana and Puigcerver, Joan and Papalampidi, Pinelopi and Henaff, Olivier and Xiong, Xi and Soricut, Radu and Harmsen, Jeremiah and Zhai, Xiaohua},
    month = jul,
    year = {2024},
}

@inproceedings{chi2024universal,
  title     = {Universal Manipulation Interface: In-The-Wild Robot Teaching Without In-The-Wild Robots},
  author    = {Chi, Cheng and Xu, Zhenjia and Pan, Chuer and Cousineau, Eric and Burchfiel, Benjamin and Feng, Siyuan and Tedrake, Russ and Song, Shuran},
  booktitle = {Proceedings of Robotics: Science and Systems (RSS)},
  year      = {2024}
}

@article{marafioti2025smolvlm,
  title={Smolvlm: Redefining small and efficient multimodal models},
  author={Marafioti, Andr{\'e}s and Zohar, Orr and Farr{\'e}, Miquel and Noyan, Merve and Bakouch, Elie and Cuenca, Pedro and Zakka, Cyril and Allal, Loubna Ben and Lozhkov, Anton and Tazi, Nouamane and others},
  journal={arXiv preprint arXiv:2504.05299},
  year={2025}
}

@article{bai2025qwen3,
  title={Qwen3-vl technical report},
  author={Bai, Shuai and Cai, Yuxuan and Chen, Ruizhe and Chen, Keqin and Chen, Xionghui and Cheng, Zesen and Deng, Lianghao and Ding, Wei and Gao, Chang and Ge, Chunjiang and others},
  journal={arXiv preprint arXiv:2511.21631},
  year={2025}
}

@inproceedings{hellaswag,
 address = {Florence, Italy},
 author = {Zellers, Rowan  and
Holtzman, Ari  and
Bisk, Yonatan  and
Farhadi, Ali  and
Choi, Yejin},
 booktitle = {Proceedings of the 57th Annual Meeting of the Association for Computational Linguistics},
 doi = {10.18653/v1/P19-1472},
 pages = {4791--4800},
 publisher = {Association for Computational Linguistics},
 title = {{H}ella{S}wag: Can a Machine Really Finish Your Sentence?},
 url = {https://aclanthology.org/P19-1472},
 year = {2019}
}

@inproceedings{mmlu,
 author = {Dan Hendrycks and
Collin Burns and
Steven Basart and
Andy Zou and
Mantas Mazeika and
Dawn Song and
Jacob Steinhardt},
 bibsource = {dblp computer science bibliography, https://dblp.org},
 booktitle = {9th International Conference on Learning Representations, {ICLR} 2021,
Virtual Event, Austria, May 3-7, 2021},
 publisher = {OpenReview.net},
 title = {Measuring Massive Multitask Language Understanding},
 url = {https://openreview.net/forum?id=d7KBjmI3GmQ},
 year = {2021}
}

@article{arc,
 author = {Peter Clark  and Isaac Cowhey and Oren Etzioni and Tushar Khot and
Ashish Sabharwal and Carissa Schoenick and Oyvind Tafjord},
 journal = {ArXiv preprint},
 title = {Think you have Solved Question Answering? Try ARC, the AI2 Reasoning Challenge},
 url = {https://arxiv.org/abs/1803.05457},
 volume = {abs/1803.05457},
 year = {2018}
}

@inproceedings{piqa,
 author = {Yonatan Bisk and
Rowan Zellers and
Ronan LeBras and
Jianfeng Gao and
Yejin Choi},
 bibsource = {dblp computer science bibliography, https://dblp.org},
 booktitle = {The Thirty-Fourth {AAAI} Conference on Artificial Intelligence, {AAAI}
2020, The Thirty-Second Innovative Applications of Artificial Intelligence
Conference, {IAAI} 2020, The Tenth {AAAI} Symposium on Educational
Advances in Artificial Intelligence, {EAAI} 2020, New York, NY, USA,
February 7-12, 2020},
 pages = {7432--7439},
 publisher = {{AAAI} Press},
 title = {{PIQA:} Reasoning about Physical Commonsense in Natural Language},
 url = {https://aaai.org/ojs/index.php/AAAI/article/view/6239},
 year = {2020}
}

@inproceedings{sakaguchi2019winogrande,
 author = {Keisuke Sakaguchi and
Ronan Le Bras and
Chandra Bhagavatula and
Yejin Choi},
 bibsource = {dblp computer science bibliography, https://dblp.org},
 booktitle = {The Thirty-Fourth {AAAI} Conference on Artificial Intelligence, {AAAI}
2020, The Thirty-Second Innovative Applications of Artificial Intelligence
Conference, {IAAI} 2020, The Tenth {AAAI} Symposium on Educational
Advances in Artificial Intelligence, {EAAI} 2020, New York, NY, USA,
February 7-12, 2020},
 pages = {8732--8740},
 publisher = {{AAAI} Press},
 title = {WinoGrande: An Adversarial Winograd Schema Challenge at Scale},
 url = {https://aaai.org/ojs/index.php/AAAI/article/view/6399},
 year = {2020}
}

@inproceedings{OpenBookQA2018,
 address = {Brussels, Belgium},
 author = {Mihaylov, Todor  and
Clark, Peter  and
Khot, Tushar  and
Sabharwal, Ashish},
 booktitle = {Proceedings of the 2018 Conference on Empirical Methods in Natural Language Processing},
 doi = {10.18653/v1/D18-1260},
 pages = {2381--2391},
 publisher = {Association for Computational Linguistics},
 title = {Can a Suit of Armor Conduct Electricity? A New Dataset for Open Book Question Answering},
 url = {https://aclanthology.org/D18-1260},
 year = {2018}
}

@inproceedings{boolq,
 address = {Minneapolis, Minnesota},
 author = {Clark, Christopher  and
Lee, Kenton  and
Chang, Ming-Wei  and
Kwiatkowski, Tom  and
Collins, Michael  and
Toutanova, Kristina},
 booktitle = {Proceedings of the 2019 Conference of the North {A}merican Chapter of the Association for Computational Linguistics: Human Language Technologies, Volume 1 (Long and Short Papers)},
 doi = {10.18653/v1/N19-1300},
 pages = {2924--2936},
 publisher = {Association for Computational Linguistics},
 title = {{B}ool{Q}: Exploring the Surprising Difficulty of Natural Yes/No Questions},
 url = {https://aclanthology.org/N19-1300},
 year = {2019}
}

@misc{lbm_eval2025,
  title        = {{LBM Eval}: A Simulation Benchmark for Large Behavior Model Policies},
  author={TRI LBM Team and Jose Barreiros and Andrew Beaulieu and Aditya Bhat and Rick Cory and Eric Cousineau and Hongkai Dai and Ching-Hsin Fang and Kunimatsu Hashimoto and Muhammad Zubair Irshad and Masha Itkina and Naveen Kuppuswamy and Kuan-Hui Lee and Katherine Liu and Dale McConachie and Ian McMahon and Haruki Nishimura and Calder Phillips-Grafflin and Charles Richter and Paarth Shah and Krishnan Srinivasan and Blake Wulfe and Chen Xu and Mengchao Zhang and Alex Alspach and Maya Angeles and Kushal Arora and Vitor Campagnolo Guizilini and Alejandro Castro and Dian Chen and Ting-Sheng Chu and Sam Creasey and Sean Curtis and Richard Denitto and Emma Dixon and Eric Dusel and Matthew Ferreira and Aimee Goncalves and Grant Gould and Damrong Guoy and Swati Gupta and Xuchen Han and Kyle Hatch and Brendan Hathaway and Allison Henry and Hillel Hochsztein and Phoebe Horgan and Shun Iwase and Donovon Jackson and Siddharth Karamcheti and Sedrick Keh and Joseph Masterjohn and Jean Mercat and Patrick Miller and Paul Mitiguy and Tony Nguyen and Jeremy Nimmer and Yuki Noguchi and Reko Ong and Aykut Onol and Owen Pfannenstiehl and Richard Poyner and Leticia Priebe Mendes Rocha and Gordon Richardson and Christopher Rodriguez and Derick Seale and Michael Sherman and Mariah Smith-Jones and David Tago and Pavel Tokmakov and Matthew Tran and Basile Van Hoorick and Igor Vasiljevic and Sergey Zakharov and Mark Zolotas and Rares Ambrus and Kerri Fetzer-Borelli and Benjamin Burchfiel and Hadas Kress-Gazit and Siyuan Feng and Stacie Ford and Russ Tedrake},
  year         = {2025},
  howpublished = {\url{https://github.com/ToyotaResearchInstitute/lbm_eval}},
  note         = {Toyota Research Institute. Version 1.1.0},
}

@article{community2026starvla,
  title={StarVLA: A Lego-like Codebase for Vision-Language-Action Model Developing},
  author={Community, StarVLA},
  journal={arXiv preprint arXiv:2604.05014},
  year={2026}
}

@article{xie2025dexbotic,
  title={Dexbotic: Open-source vision-language-action toolbox},
  author={Xie, Bin and Zhou, Erjin and Jia, Fan and Shi, Hao and Fan, Haoqiang and Zhang, Haowei and Li, Hebei and Sun, Jianjian and Bin, Jie and Huang, Junwen and others},
  journal={arXiv preprint arXiv:2510.23511},
  year={2025}
}

@misc{pfaff2025drakeblendertools,
  author       = {Pfaff, Nicholas and Werner, Peter},
  title        = {Drake Blender Tools: Importing {Drake} Simulations into {Blender}},
  year         = {2025},
  howpublished = {\url{https://github.com/nepfaff/drake-blender-tools}},
}

@article{bandaru2025foundation,
  title   = {Foundation Models for Robotics: Vision-Language-Action (VLA)},
  author  = {Bandaru, Rohit},
  year    = {2025},
  month   = {Sep},
  url     = {https://rohitbandaru.github.io/blog/Foundation-Models-for-Robotics-VLA/}
}

@article{liu2023llm360,
  title   = {{LLM360}: Towards Fully Transparent Open-Source {LLMs}},
  author  = {Liu, Zhengzhong and Qiao, Aurick and Neiswanger, Willie and Wang, Hongyi and Tan, Bowen and Tao, Tianhua and Li, Junbo and Wang, Yuqi and Sun, Suqi and Pangarkar, Omkar and Fan, Richard and Gu, Yi and Miller, Victor and Zhuang, Yonghao and He, Guowei and Li, Haonan and Koto, Fajri and Tang, Liping and Ranjan, Nikhil and Shen, Zhiqiang and Ren, Xuguang and Iriondo, Roberto and Mu, Cun and Hu, Zhiting and Schulze, Mark and Nakov, Preslav and Baldwin, Timothy and Xing, Eric P.},
  journal = {arXiv preprint arXiv:2312.06550},
  year    = {2023},
  doi     = {10.48550/arXiv.2312.06550},
  url     = {https://arxiv.org/abs/2312.06550}
}

@article{liu2025k2,
  title   = {{LLM360 K2}: Building a 65{B} 360-Open-Source Large Language Model from Scratch},
  author  = {Liu, Zhengzhong and Tan, Bowen and Wang, Hongyi and Neiswanger, Willie and Tao, Tianhua and Li, Haonan and Koto, Fajri and Wang, Xiaodan and Sun, Suqi and Pangarkar, Omkar and Fan, Richard and Gu, Yi and Miller, Victor and Zhuang, Yonghao and He, Guowei and Tang, Liping and Ranjan, Nikhil and Shen, Zhiqiang and Ren, Xuguang and Iriondo, Roberto and Mu, Cun and Hu, Zhiting and Schulze, Mark and Nakov, Preslav and Baldwin, Timothy and Xing, Eric P.},
  journal = {arXiv preprint arXiv:2501.07124},
  year    = {2025},
  doi     = {10.48550/arXiv.2501.07124},
  url     = {https://arxiv.org/abs/2501.07124}
}

@software{fastllm2024,
  title        = {Fast-{LLM}: Accelerating Your {LLM} Training to Full Speed},
  author       = {{ServiceNow Research}},
  year         = {2024},
  publisher    = {GitHub},
  url          = {https://github.com/ServiceNow/Fast-LLM},
  license      = {Apache-2.0}
}

@software{karpathy2022nanogpt,
  title     = {nano{GPT}: The Simplest, Fastest Repository for Training/Finetuning Medium-Sized {GPT}s},
  author    = {Karpathy, Andrej},
  year      = {2022},
  publisher = {GitHub},
  url       = {https://github.com/karpathy/nanoGPT},
  license   = {MIT}
}

@book{build-llms-from-scratch-book,
  author       = {Sebastian Raschka},
  title        = {Build A Large Language Model (From Scratch)},
  publisher    = {Manning},
  year         = {2024},
  isbn         = {978-1633437166},
  url          = {https://www.manning.com/books/build-a-large-language-model-from-scratch},
  github       = {https://github.com/rasbt/LLMs-from-scratch}
}

@inproceedings{li2023blip2,
  title     = {{BLIP-2}: Bootstrapping Language-Image Pre-training with Frozen Image Encoders and Large Language Models},
  author    = {Li, Junnan and Li, Dongxu and Savarese, Silvio and Hoi, Steven},
  booktitle = {Proceedings of the 40th International Conference on Machine Learning (ICML)},
  year      = {2023},
  url       = {https://arxiv.org/abs/2301.12597}
}

@inproceedings{rasley2020deepspeed,
  title     = {{DeepSpeed}: System Optimizations Enable Training Deep Learning Models with Over 100 Billion Parameters},
  author    = {Rasley, Jeff and Rajbhandari, Samyam and Ruwase, Olatunji and He, Yuxiong},
  booktitle = {Proceedings of the 26th ACM SIGKDD International Conference on Knowledge Discovery \& Data Mining},
  pages     = {3505--3506},
  year      = {2020},
  publisher = {ACM},
  doi       = {10.1145/3394486.3406703},
  url       = {https://github.com/deepspeedai/DeepSpeed}
}

@software{breuel2020webdataset,
  title     = {{WebDataset}: A High-Performance Python-Based {I/O} System for Large Deep Learning Problems},
  author    = {Breuel, Thomas},
  year      = {2020},
  publisher = {GitHub},
  url       = {https://github.com/webdataset/webdataset},
  license   = {BSD-3-Clause}
}

@inproceedings{moritz2018ray,
  title     = {Ray: A Distributed Framework for Emerging {AI} Applications},
  author    = {Moritz, Philipp and Nishihara, Robert and Wang, Stephanie and Tumanov, Alexey and Liaw, Richard and Liang, Eric and Elibol, Melih and Yang, Zongheng and Paul, William and Jordan, Michael I. and Stoica, Ion},
  booktitle = {Proceedings of the 13th USENIX Symposium on Operating Systems Design and Implementation (OSDI)},
  pages     = {561--577},
  year      = {2018},
  url       = {https://www.usenix.org/conference/osdi18/presentation/moritz}
}

@article{kaplan2020scaling,
  title={Scaling laws for neural language models},
  author={Kaplan, Jared and McCandlish, Sam and Henighan, Tom and Brown, Tom B and Chess, Benjamin and Child, Rewon and Gray, Scott and Radford, Alec and Wu, Jeffrey and Amodei, Dario},
  journal={arXiv preprint arXiv:2001.08361},
  year={2020}
}

@article{li2024datacomp,
  title={Datacomp-lm: In search of the next generation of training sets for language models},
  author={Li, Jeffrey and Fang, Alex and Smyrnis, Georgios and Ivgi, Maor and Jordan, Matt and Gadre, Samir and Bansal, Hritik and Guha, Etash and Keh, Sedrick and Arora, Kushal and others},
  journal={Advances in Neural Information Processing Systems},
  volume={37},
  pages={14200--14282},
  year={2024}
}

@article{lipman2022flow,
  title={Flow matching for generative modeling},
  author={Lipman, Yaron and Chen, Ricky TQ and Ben-Hamu, Heli and Nickel, Maximilian and Le, Matt},
  journal={arXiv preprint arXiv:2210.02747},
  year={2022}
}

@inproceedings{shi2016real,
  title={Real-time single image and video super-resolution using an efficient sub-pixel convolutional neural network},
  author={Shi, Wenzhe and Caballero, Jose and Husz{\'a}r, Ferenc and Totz, Johannes and Aitken, Andrew P and Bishop, Rob and Rueckert, Daniel and Wang, Zehan},
  booktitle={Proceedings of the IEEE conference on computer vision and pattern recognition},
  pages={1874--1883},
  year={2016}
}

@article{chen2015microsoft,
  title={Microsoft coco captions: Data collection and evaluation server},
  author={Chen, Xinlei and Fang, Hao and Lin, Tsung-Yi and Vedantam, Ramakrishna and Gupta, Saurabh and Doll{\'a}r, Piotr and Zitnick, C Lawrence},
  journal={arXiv preprint arXiv:1504.00325},
  year={2015}
}

@software{gpt-neox-library,
  title = {{GPT-NeoX: Large Scale Autoregressive Language Modeling in PyTorch}},
  author = {Andonian, Alex and Anthony, Quentin and Biderman, Stella and Black, Sid and Gali, Preetham and Gao, Leo and Hallahan, Eric and Levy-Kramer, Josh and Leahy, Connor and Nestler, Lucas and Parker, Kip and Pieler, Michael and Phang, Jason and Purohit, Shivanshu and Schoelkopf, Hailey and Stander, Dashiell and Songz, Tri and Tigges, Curt and Thérien, Benjamin and Wang, Phil and Weinbach, Samuel},
  url = {https://www.github.com/eleutherai/gpt-neox},
  doi = {10.5281/zenodo.5879544},
  month = {9},
  year = {2023},
  version = {2.0.0},
}

@article{penedo2024fineweb,
  title={The fineweb datasets: Decanting the web for the finest text data at scale},
  author={Penedo, Guilherme and Kydl{\'\i}{\v{c}}ek, Hynek and Lozhkov, Anton and Mitchell, Margaret and Raffel, Colin and Von Werra, Leandro and Wolf, Thomas and others},
  journal={Advances in Neural Information Processing Systems},
  volume={37},
  pages={30811--30849},
  year={2024}
}

@inproceedings{radford2021learning,
  title={Learning transferable visual models from natural language supervision},
  author={Radford, Alec and Kim, Jong Wook and Hallacy, Chris and Ramesh, Aditya and Goh, Gabriel and Agarwal, Sandhini and Sastry, Girish and Askell, Amanda and Mishkin, Pamela and Clark, Jack and others},
  booktitle={International conference on machine learning},
  pages={8748--8763},
  year={2021},
  organization={PmLR}
}

@misc{drake,
  author = {Russ Tedrake and the Drake Development Team},
  title = {Drake: Model-based design and verification for robotics},
  year = {2019},
  url = {https://drake.mit.edu}
}


\beginappendix

\section{VLA Foundry -- Detailed Reference}
\label{app:foundry-details}

This appendix expands on the description of VLA Foundry in Section~\ref{sec:foundry}. Section~\ref{app:framework} covers the core framework internals (configuration system, registry, dataloading, dataset mixing, and preprocessing) with representative code and configuration snippets alongside each description, and Section~\ref{app:robotics} covers robotics-specific utilities (normalization, action representations, sampling windows, and proprioception).

\subsection{Framework Internals}
\label{app:framework}

\subsubsection{Model Training Loop}
\label{app:model-training}
Training in VLA Foundry is done in a single training loop that is shared across every model stage in the pipeline: LLM pretraining, VLM pretraining, and VLA fine-tuning. VLA Foundry takes a data-centered approach. It expresses training budgets as a number of samples instead of a number of steps, so that runs at different batch sizes or GPU counts remain directly comparable.

The training loop is deliberately model-agnostic. Unpacking a batch into inputs and targets, masking the loss, and reducing model outputs to a scalar is attached to each model class rather than baked into the loop, so the same training pathway drives an LLM learning from raw text, a VLM learning from image--caption pairs, and a VLA learning to denoise actions. The loop composes cleanly with the distributed training primitives users expect: FSDP with optional CPU offloading, mixed-precision execution, gradient accumulation, gradient checkpointing, \texttt{torch.compile}, and exponential moving average of weights.

\subsubsection{Config System and Argument Parsing}
\label{app:config-system}
We use \texttt{draccus}~\cite{draccus} to parse arguments. At the most basic level, arguments are supplied as command-line flags. Optionally, to avoid manually typing many flags, users can supply \texttt{--config\_path} and point to a YAML file with nested parameters, and YAML files themselves can be nested with the \texttt{include} keyword.

Configurations follow a three-level precedence: command-line arguments override YAML preset files, which override dataclass defaults. Parameters are organized hierarchically and any field can be overridden at arbitrary nesting depth. To give a concrete example, consider the command \texttt{python main.py --config\_path config.yaml --model.hidden\_dim=1024}. Here, the order of priority would be (1) the CLI flag \texttt{--model.hidden\_dim=1024}, (2) the \texttt{hidden\_dim} in \texttt{config.yaml}, (3) [if it exists] the \texttt{hidden\_dim} in any nested \texttt{include} file, (4) the default values defined within the code.

A \texttt{--resolve\_configs} flag prints the complete merged configuration and generates a YAML, letting users verify exactly what will run. A minimal example of a nested YAML configuration, invoked with \texttt{python main.py --config\_path config.yaml}, is shown below.

\begin{lstlisting}[style=codebase, language=yaml, backgroundcolor=\color{gray!15}]
# config.yaml
  model:
    include: vla_foundry/config_presets/models/transformer_11m.yaml
    hidden_dim: 2048  # overrides the preset value
    vit:
      include: vla_foundry/config_presets/models/vit_paligemma.yaml
  hparams:
    lr: 1e-4
    global_batch_size: 256
    per_gpu_batch_size: 8
    precision: amp_bfloat16
\end{lstlisting}
\label{app:config-example}

\subsubsection{Dataset and Model Registry}
\label{app:registry}
The \texttt{--model} and \texttt{--dataset} arguments have a special keyword called \texttt{type}. The \texttt{--model.type} and \texttt{--data.type} arguments select which parameter class and model or data pipeline to instantiate at runtime. Each model is registered via a \texttt{@register\_model} decorator on its factory function, and the same pattern applies to datasets.

This means that adding a new model to VLA Foundry requires only two things: a frozen dataclass defining its hyperparameters, and a factory function decorated with \texttt{@register\_model}. No central configuration file needs to be modified. At runtime, \texttt{create\_model()} looks up the registry by \texttt{model.type} and dispatches to the correct factory. Each model selects a \texttt{BatchHandler} -- typically one of the shared handlers defined per modality or model type (LLM, VLM, and DP-VLA) -- which encapsulates batching, loss construction, and output reduction, and keeps the main training loop model-agnostic. Entirely new training paradigms can register an additional handler via \texttt{@register\_batch\_handler}. Registering a new model looks like:

\begin{lstlisting}[style=codebase, language=python, backgroundcolor=\color{gray!15}]
# This can be accessed with `--model.type = diffusion_policy`
  class DiffusionPolicy(self, [...]):
     # Implementation here

  @register_model("diffusion_policy")
  def create_diffusion_policy(model_params: ModelParams, load_pretrained: bool = True):
      vision_language_backbone = get_vision_language_backbone(
          model_params.vision_language_backbone, load_pretrained
      )
      transformer = create_model(model_params.transformer, load_pretrained)
      noise_scheduler = create_noise_scheduler(model_params)
      return DiffusionPolicy(model_params, vision_language_backbone, transformer, noise_scheduler)
\end{lstlisting}
\label{app:registry-example}

\subsubsection{Dataloading}
\label{app:dataloading}
We use WebDatasets for dataloading and store the data in tar shards. Within each tar file, each sample is distinguished by its unique prefix. The structure of the directory is shown below. This structure is designed to be extensible, where new fields can be added easily if necessary. For instance, if we want to include depth images, we can do this by adding \texttt{unique\_name\_1\_depth1.jpg}. The flexibility of our data format also allows us to extend to other modalities such as video.

\dirtree{%
  .1 dataset\_name/.
  .2 manifest.jsonl.
  .2 shard\_00000000.tar.
  .3 unique\_name\_1\_camera1.jpg.
  .3 unique\_name\_1\_camera2.jpg.
  .3 unique\_name\_1\_meta.json.
  .3 unique\_name\_1\_actions.npz.
  .3 unique\_name\_2\_camera1.jpg.
  .3 \ldots.
  .2 shard\_00000001.tar.
  .2 shard\_00000002.tar.
  .2 \ldots.
}

The data processing pipeline steps are defined separately for each modality. Currently, the data modalities supported are text, caption (text+image), and robotics (text+image+action). The steps of the WebDataset pipeline are defined sequentially, and notably support both WebDataset built-in functions (e.g., \texttt{wds.split\_by\_node}) and user-defined functions that can be composed freely. This is especially useful for the robotics processing detailed in Section~\ref{app:robotics}. An example pipeline for an image-caption dataset follows; the same composition pattern is used for text and robotics pipelines, with different per-modality steps.

\begin{lstlisting}[style=codebase, language=python, backgroundcolor=\color{gray!15}]
pipeline = [
    wds.SimpleShardList(datastring),
    deterministic_shuffle(
        bufsize=self.data_params.shuffle_buffer_size,
        initial=self.data_params.shuffle_initial,
        seed=self.data_params.seed,
        epoch=checkpoint_num,
    ),
    wds.split_by_node,
    wds.split_by_worker,
    wds.tarfile_to_samples(handler=log_and_continue),
    wds.decode("pilrgb", handler=log_and_continue),
    wds.select(filter_no_caption_or_no_image),
    wds.map(
        lambda sample: self.augmentations.apply_transforms(sample),
        handler=log_and_continue,
    ),
    wds.rename(image="jpg;png;jpeg;webp", text="txt"),
    wds.map(lambda sample: {**sample, "text": "<image> " + sample["text"]}),
    wds.batched(self.batch_size, partial=False),
    wds.map(
        lambda sample: self.processor(
            images=sample["image"],
            text=sample["text"],
            return_tensors="pt",
            padding="max_length",
            padding_side="right",
            max_length=self.data_params.seq_len + 1,
        ),
        handler=log_and_continue,
    ),
    wds.map(
        lambda sample: {
            "input_ids": sample["input_ids"],
            "attention_mask": sample["attention_mask"],
            "pixel_values": sample["pixel_values"],
        }
    ),
]
return pipeline
\end{lstlisting}
\label{app:dataloading-example}

\subsubsection{Dataset Mixing}
\label{app:dataset-mixing}
VLA Foundry natively supports dataset mixing with command-line arguments. By default, the dataset-related arguments are lists, which means that supporting multiple datasets is as simple as adding elements to the list. Of special note is the \texttt{--data.dataset\_weighting} parameter, which handles the batch balancing ratios; a 1:2:1 weighting corresponds to 25\%/50\%/25\% of each batch drawn from the respective datasets. An example YAML snippet that mixes three robotics datasets with a 1:2:1 weighting is shown below.

\begin{lstlisting}[style=codebase, language=yaml, backgroundcolor=\color{gray!15}]

  dataset_manifest:
    - tasks_processed/BimanualPlaceAppleFromBowlIntoBin/shards/manifest.jsonl
    - tasks_processed/BimanualPlaceFruitFromBowlIntoBin/shards/manifest.jsonl
    - tasks_processed/BimanualPutRedBellPepperInBin/shards/manifest.jsonl
  dataset_statistics:
    - tasks_processed/BimanualPlaceAppleFromBowlIntoBin/shards/stats.json
    - tasks_processed/BimanualPlaceFruitFromBowlIntoBin/shards/stats.json
    - tasks_processed/BimanualPutRedBellPepperInBin/shards/stats.json
  dataset_modality:
    - robotics
    - robotics
    - robotics
  dataset_weighting:
    - 1.0
    - 2.0
    - 1.0
\end{lstlisting}
\label{app:dataset-mixing-example}

\subsubsection{Preprocessing and Manifests}
\label{app:preprocessing}
VLA Foundry has custom scripts to convert raw datasets to the WebDataset tar shards described above. As noted in Section~\ref{app:dataloading}, we currently support text, image-caption, and robotics datasets. Text preprocessing reads parquet files (typically stored on S3) and emits one JSON sample per row. Image-caption preprocessing takes a URL list and downloads image--text pairs via \texttt{img2dataset}. Utilities are also shipped for fetching upstream data from Hugging Face Hub and from HTTP directory listings into the intermediate storage consumed by these scripts. Robotics raw data can come from any source (simulation logs, real-robot recordings, etc.), as long as a converter knows how to read it and produce a standardized output; for this release, we provide converters from LeRobot and from the Spartan format used in \texttt{lbm\_eval}. These robotics converters all share the same entry point and follow the same logic, so adding a new one amounts to creating a new class that inherits from \texttt{BaseRoboticsConverter} and filling in the necessary methods such as \texttt{discover\_cameras}.

We use \texttt{ray}~\cite{moritz2018ray} to parallelize data preprocessing. Under the \texttt{ray} framework, there is a head node that orchestrates the jobs and several worker nodes that each run a small independent job. For robotics datasets, this is done in multiple stages. First, it creates a \texttt{frames} folder in the output directory, where each sample is its own unique tar file. Next, it creates an \texttt{episodes} folder, where the sample tar files are grouped together by episode. Finally, it creates a \texttt{shards} folder, where sample tar files are grouped together randomly. This \texttt{shards} folder is what is ultimately used for training. Within the \texttt{shards} folder, there is a \texttt{manifest.json} which contains an overview of the shards; an example follows.

\begin{lstlisting}[style=codebase, language=json, backgroundcolor=\color{gray!15}]
{"shard": "00000000", "num_sequences": 1024}
{"shard": "00000001", "num_sequences": 1024}
{"shard": "00000002", "num_sequences": 1024}
{"shard": "00000003", "num_sequences": 488}
\end{lstlisting}
\label{app:manifest-example}

Robotics datasets additionally have a \texttt{stats.json} within the \texttt{shards} folder, which contains statistics computed across all samples of the dataset. This computation requires worker nodes to first store local statistics in node memory, then communicate and gather across different nodes. For internal runs, this was tested on AWS EC2 with 60 nodes of \texttt{i4i.4xlarge}, but we have tested it locally as well.

\subsection{Robotics-specific Details}
\label{app:robotics}

\subsubsection{Normalization}
\label{app:normalization}
Actions and proprioceptive states are normalized during dataloading time and denormalized at inference time. Normalization is handled by a \texttt{RoboticsNormalizer} class, which supports four normalization methods: standard deviation, min-max, and two percentile-based variants (\texttt{percentile\_1\_99} and \texttt{percentile\_5\_95}). The choice of percentile-based normalization is useful for action fields that contain outliers, as it avoids compressing the bulk of the distribution to a narrow band. Statistics are precomputed across the full dataset during preprocessing and stored in a \texttt{stats.json} file alongside each dataset shard.

\paragraph{\textbf{Normalization scope}} Normalization can be applied at two scopes. In \emph{global} scope, the scalars \texttt{mean} and \texttt{scale} are applied uniformly across all timesteps in a sequence. In \emph{per-timestep} scope, each timestep within the action window has its own mean and scale derived from statistics computed at that relative offset in the trajectory. Per-timestep normalization is particularly useful for relative action representations, where the distribution of predicted displacements can vary considerably between early and late steps of the prediction horizon. When working with cropped sequences (see Section~\ref{app:action-window}), per-timestep statistics are aligned to the anchor timestep so that indices into the statistics tensor correspond correctly to the tensor's time axis.

\paragraph{\textbf{Merging statistics}} When training on multiple datasets simultaneously (Section~\ref{app:dataset-mixing}), users may wish to use the joint distribution across all datasets rather than any individual one. Since datasets are processed individually with their own per-dataset \texttt{stats.json} files, we support merging multiple \texttt{stats.json} files together. Means are computed as sample-count-weighted averages. Standard deviations are merged using the law of total variance, $\sigma^2_{\text{overall}} = \mathbb{E}[\sigma^2_i]$. Min and max statistics are obtained as element-wise minima and maxima across datasets. Percentiles cannot be merged exactly from summary statistics alone; instead, each dataset retains a serialized t-digest sketch~\cite{dunning2019computing} during preprocessing, and the sketches are merged at training time to recover approximate percentiles of the pooled distribution. All statistics are computed and merged per action-space dimension, and optionally per timestep within the prediction window when using per-timestep normalization scope.

\subsubsection{Absolute vs.\ Relative Actions}
\label{app:absolute-relative}
VLA Foundry supports both absolute and relative action representations, which are stored as separate fields in the dataset. Absolute actions are poses expressed in the world frame (e.g., end-effector XYZ position and 6D rotation). Relative actions are computed with respect to the robot's actual pose at the anchor timestep, i.e., the frame at which a prediction is made.

Formally, let $T_{\text{ref}} \in SE(3)$ denote the actual end-effector pose at the anchor timestep and $T_t \in SE(3)$ the action pose at future timestep $t$. The relative action is defined as
\[
T_t^{\text{rel}} = T_{\text{ref}}^{-1} \cdot T_t,
\]
where the product is the standard $SE(3)$ group operation. Rotations are represented in the 6D continuous rotation format~\cite{zhou2019continuity} throughout, with conversion to and from $SO(3)$ matrices performed via Gram--Schmidt orthogonalization. VLA Foundry's preprocessing scripts generate both absolute and relative fields given configurations defining which fields form poses, and the practitioner selects which to use via the \texttt{--action\_fields} configuration during training.

\subsubsection{Past/Future Action Window}
\label{app:action-window}
During dataset preprocessing, each training sample is constructed around an \emph{anchor timestep} $t$ within an episode. The low-dimensional window centered at $t$ spans $[t - N_{\text{past}},\, t + N_{\text{future}}]$, where $N_{\text{past}}$ and $N_{\text{future}}$ are configurable preprocessing parameters (\texttt{past\_lowdim\_steps} and \texttt{future\_lowdim\_steps}). This produces a tensor of $N_{\text{past}} + 1 + N_{\text{future}}$ timesteps per sample. Including past timesteps allows the model to condition on recent action history and proprioceptive context; predicting multiple future timesteps allows for temporal action chunking~\cite{Zhao-RSS-23}.

At episode boundaries, sequences are padded using a configurable padding strategy (copy, zero, or reflect). To avoid degenerate samples with excessive padding, samples whose required padding exceeds configurable thresholds (\texttt{max\_padding\_left}, \texttt{max\_padding\_right}) are discarded during preprocessing. The anchor timestep's position within the cropped window is stored in sample metadata as \texttt{anchor\_relative\_idx}, enabling downstream code to correctly align per-timestep normalization statistics and to separate past from future timesteps without re-parsing raw episode indices. Notably, the preprocessing past/future action window does not need to be identical to the past/future values used during training. This allows users to specify a larger window during preprocessing time, then work with a truncated subwindow during training.

\subsubsection{Proprioception}
\label{app:proprioception}
Proprioceptive state is specified via a separate \texttt{--proprioception\_fields} parameter, distinct from \texttt{--action\_fields}. Typical proprioception fields include joint positions, joint velocities, and actual end-effector poses (XYZ and 6D rotation). During batch construction, the fields listed in \texttt{proprioception\_fields} are each extracted, normalized, and concatenated along the feature dimension to form a single \texttt{proprioception} tensor of shape $[B,\, T_{\text{prop}},\, D_{\text{prop}}]$. A key design difference from actions is that proprioception uses only the \emph{past and current} timesteps within the window (i.e., indices $[0,\, t_{\text{anchor}}]$), whereas actions span the full past-and-future window. This reflects the causal structure of the problem: past proprioception is observed history available to the policy, while future proprioception is not available at inference time.

\section{Links to Checkpoints and Additional Resources} \label{sec:links}

\textbf{Project website:} \foundryWebsite{}

\textbf{Project code:} \foundryCodebase{}

\textbf{Model weights:} \hfmodels{}

\section{LLM-VLM-VLA Details}

\subsection{Model Sizes}

Table \ref{tab:model-sizes} details the different module sizes of the two architectures used in this report.

  \begin{table}[h]
  \centering                               
  \caption{Parameter count (billions). Embedding = VLM input embedding + output projection (lm\_head) + ViT patch/position embed. Non-embed = LLM + Vision + Action head.}
  \label{tab:model-sizes}
  \begin{tabular}{lcccccc}
  \toprule
  Model & Embedding & LLM & Vision & Action head & Total & Non-embed \\
  \midrule
  \foundryVLA{}     & 0.20 & 1.23 & 0.09 & 0.33 & 1.85 & 1.65 \\
  \foundryQwenVLA{}  & 0.62 & 1.41 & 0.41 & 0.31 & 2.75 & 2.13 \\
  \bottomrule
  \end{tabular}
  \end{table}     

\subsection{LLM Benchmarks}

The following short descriptions of the benchmarks below are borrowed from \cite{li2024datacomp}.
\label{app:llm_bench}
\begin{itemize}
\item HellaSwag~\cite{hellaswag} (10,042 examples) is a 4-way multiple choice
commonsense reasoning dataset, where the model is required to understand implicit context
and common knowledge in order to correctly select the continuation to a context. HellaSwag is distributed under the MIT license as indicated in \url{https://github.com/rowanz/hellaswag/blob/master/LICENSE}.
\item MMLU~\cite{mmlu} (14,042 examples) is a 4-way multiple choice question
answering dataset that covers 57 different domains and tasks, evaluating both world
knowledge and problem solving capabilities. MMLU is distributed under the MIT license as indicated in \url{https://github.com/hendrycks/test/blob/master/LICENSE}.
\item The ARC easy (2,376 examples) and ARC challenge (1,172 examples) datasets~\cite{arc}
contain four-way multiple choice questions taken from grade 3-9 science exams, where
questions in the easy dataset require knowledge of basic science, and the challenge questions
require some procedural reasoning. are distributed under the Creative Commons Attribution-Sharealike 4.0 International license as indicated in \url{https://allenai.org/data/arc}.
\item PIQA~\cite{piqa} (1,838 examples) is a binary multiple choice question answering dataset
that requires the model to use physical commonsense reasoning to answer correctly. PIQA is distributed under the \href{https://opensource.org/licenses/AFL-3.0}{Academic Free License v. 3.0} as indicated in \url{https://github.com/ybisk/ybisk.github.io/tree/master/piqa}.
\item The Winogrande~\cite{sakaguchi2019winogrande} (273 examples) is binary multiple choice
pronoun resolution task where the model is given a context and asked to determine which
entity a pronoun refers to, requiring the model to exhibit commonsense knowledge and
contextual understanding. Winogrande is distributed under the Apache 2.0 license as indicated in \url{https://github.com/allenai/winogrande/blob/master/LICENSE}.
\item OpenBookQA~\cite{OpenBookQA2018} (500 examples) is a 4-way multiple choice question answering
dataset that requires the model to use multi-step reasoning and commonsense knowledge. OpenBookQA is distributed under the Apache 2.0 license as indicated in \url{https://github.com/allenai/OpenBookQA/blob/main/LICENSE}.
\item BoolQ~\cite{boolq} (3,270 examples) is a binary question answering dataset where the
model is expected to answer questions about relevant passages. BoolQ is distributed under the Creative Commons Share-Alike 3.0 license as indicated in \url{https://huggingface.co/datasets/google/boolq}.
\end{itemize}

\subsection{VLM Benchmark}
\label{app:vlm_bench}

COCO Captions~\cite{chen2015microsoft} (5,000 validation examples) is an image captioning dataset where the model is given an image and is required to generate a natural language description capturing the salient objects, actions, and scene context. Each image is paired with five human-written reference captions, and model outputs are evaluated using standard metrics such as CIDEr and BLEU. COCO Captions annotations are distributed under the Creative Commons Attribution 4.0 International license as indicated in \url{https://cocodataset.org/#termsofuse}. The images retain their original Flickr licenses, and use of the images must abide by the Flickr Terms of Use.

\subsection{Image Encoding Details}
\label{app:pixel_shuffle}

\begin{figure}[H]
    \centering
    \includegraphics[width=0.8\linewidth, trim= 4cm 4cm 4cm 4cm, clip]{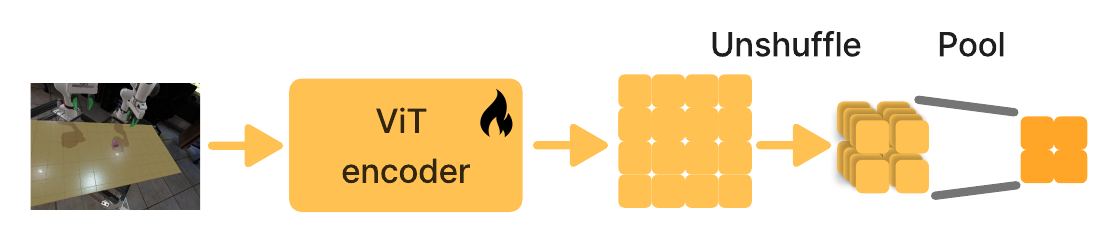}
    \caption{Representation of the pixel-shuffle operation~\cite{marafioti2025smolvlm} used for patch pooling, reducing the number of tokens passed to the downstream VLM}
    \label{fig:pixel_shuffle}
\end{figure}

\begin{figure}[h]
    \centering
    \includegraphics[width=\textwidth]{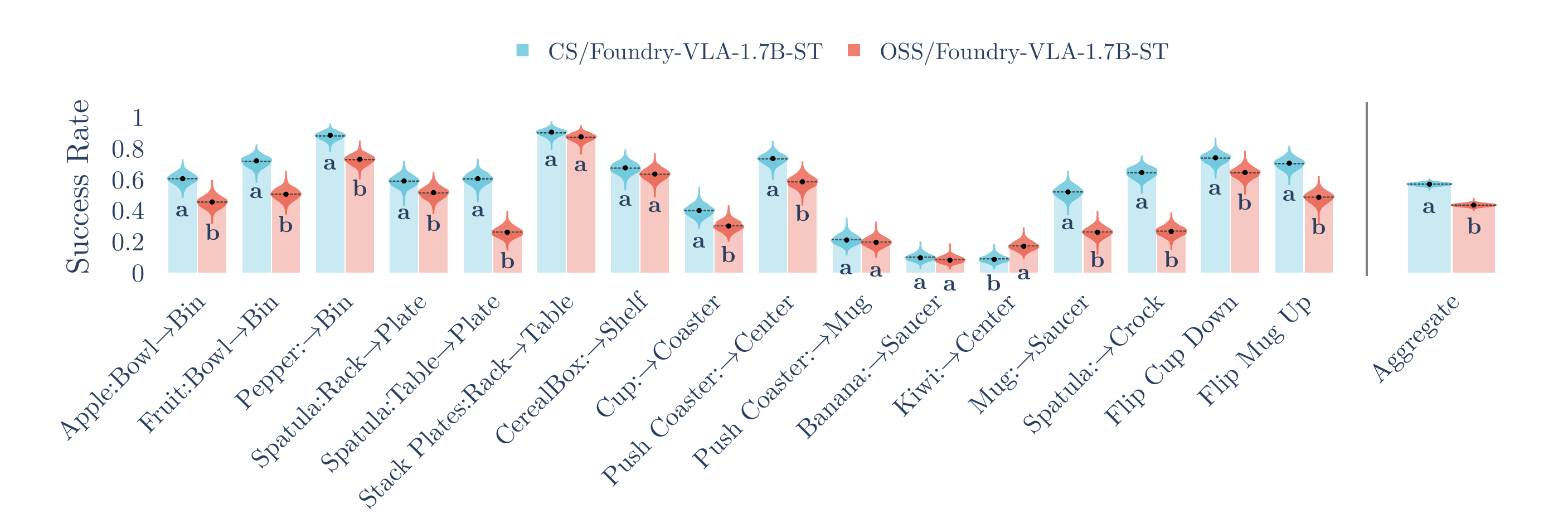}
    \vspace{0.5em}
    \includegraphics[width=\textwidth]{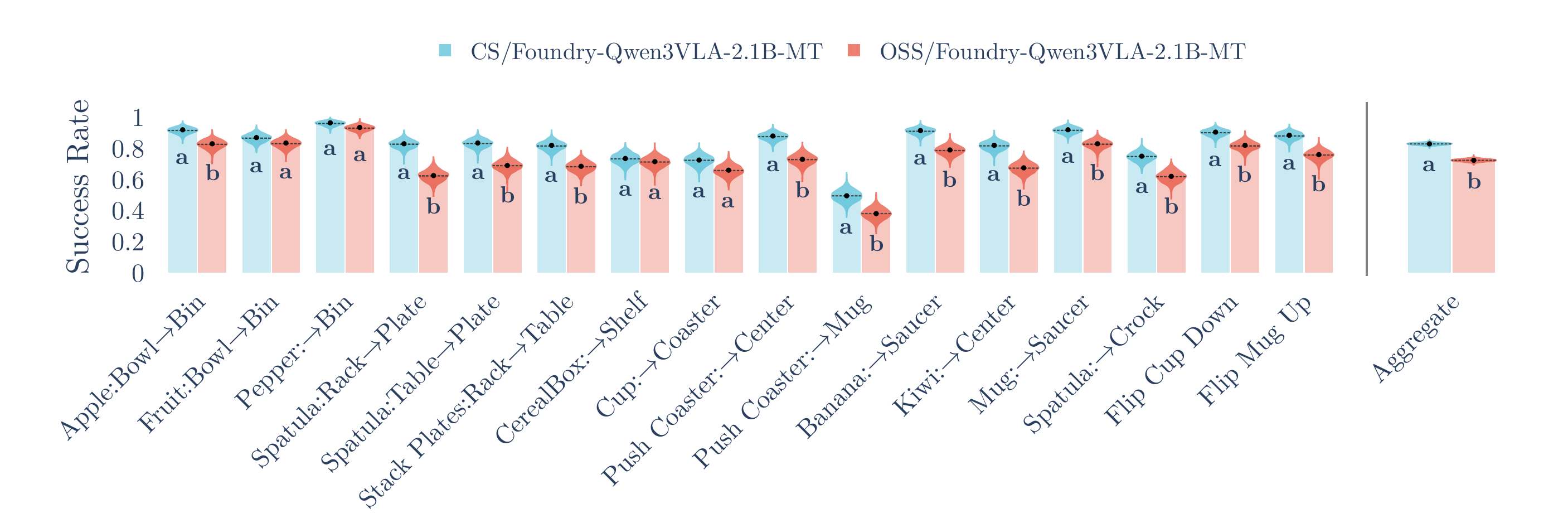}
    \caption{Comparison of checkpoints on \lbmevaloss{} (OSS) and \lbmevalcs{} (CS). In aggregate, the performance of both the \foundryVLA{} single task checkpoints and the \foundryQwenVLA{} multi task checkpoint is weaker on the open source version of the benchmark, which can be considered a distribution shifted version of the closed source version.}
    \label{fig:cs_vs_os_sim_results}
\end{figure}

Figure~\ref{fig:pixel_shuffle} shows the image encoding operation with an explicit representation of the "pixel-shuffle" pooling operation. Note that "pixel-shuffle" is usually the opposite operation~\cite{shi2016real} used for super-resolution. We label it "unshuffle" in the figure for clarity.

\subsection{Training Parameters}
\label{app:training}

Table~\ref{tab:training-hparams} shows the main training parameters used to train our different VLA models.

\begin{table}[H] 
  \centering
  \small
  \begin{tabular}{lrlrrr}
  \toprule
  Model & LR & Schedule & Warmup & Total samples & Batch size \\
  \midrule
  Foundry-VLA-1.7B-full       & $5\times10^{-5}$ & cosine & 1{,}000 & 102{,}400{,}000 & 1{,}024 \\
  Foundry-VLA-1.7B-sim        & $5\times10^{-5}$ & cosine & 1{,}000 & 102{,}400{,}000 & 1{,}024 \\
  Foundry-VLA-1.7B-real       & $5\times10^{-5}$ & cosine & 1{,}000 & 102{,}400{,}000 & 1{,}024 \\
  Foundry-VLA-1.7B-ST         & $5\times10^{-5}$ & cosine & 1{,}000 &   5{,}120{,}000 &    512 \\
  Foundry-VLA-1.7B-FT         & $5\times10^{-6}$ & cosine & 1{,}000 &   1{,}024{,}000 &    512 \\
  Foundry-VLA-1.7B-FT-sim     & $5\times10^{-6}$ & cosine & 1{,}000 &   1{,}024{,}000 &    512 \\
  Foundry-Qwen3VLA-2.1B       & $5\times10^{-5}$ & cosine & 1{,}000 & 100{,}000{,}000 & 1{,}024 \\ 
  Foundry-Qwen3VLA-2.1B-ST    & $5\times10^{-5}$ & cosine & 1{,}000 &   2{,}000{,}000 &    512 \\
  Foundry-Qwen3VLA-2.1B-FT    & $5\times10^{-6}$ & cosine & 1{,}000 &   1{,}024{,}000 &    512 \\
  \bottomrule
  \end{tabular}
  \caption{Training hyperparameters for Foundry VLA model variants. All models use AdamW with cosine learning-rate schedule and 1{,}000 warmup steps. MT variants train for $\sim$100M samples at batch 1{,}024; per-task ST trains for 2--5M samples at batch 512; per-task FT      
  fine-tunes from the MT checkpoint for 1M samples at 10$\times$ lower LR.} 
  \label{tab:training-hparams}
  \end{table}    

\subsection{VLA Dataset Details}\label{sec:vla_dataset}
As shown in Tables \ref{tab:data_overview} and \ref{tab:data_per_task_breakdown}, our subset of simulation and real data differs from the training split used in \cite{lbmtri2025} to train \lbm{}. While a small number of episodes were dropped during pre-processing, the overall dataset size is slightly larger primarily due to differences in filtering criteria and the inclusion of data previously reserved for validation. Of the internal real and simulated data, the \lbm{} models are trained on the data under column LBM; all other models are trained on the data under column VLA Foundry. Importantly, the multi-task pretrained \lbm{} model is trained on a larger dataset which includes open source OXE \cite{open_x_embodiment_rt_x_2023} data; refer to \cite{lbmtri2025} for further details. Table \ref{tab:data_manifest} shows the number of training samples per dataset split used to train VLA Foundry models; the number of samples generated by a single demonstration episode depends on the length of each demonstration and preprocessing configurations such as padding. While the internally collected real and simulated data is largely shared between \foundryVLA{}, \foundryQwenVLA{}, and \lbm{}, the VLA Foundry models use substantially more finetuning data on the unseen tasks compared to the single task and finetuned versions of \lbm{}, which we do not compare to in this technical report. Instructions on how to download the tar files used to train the sim data only variants of VLA Foundry models can be found in the released codebase.

\begin{table}[h]
\centering
\caption{Dataset overview. Previous work incorrectly categorized the ``PushBox'' simulation task as a real task.}
\label{tab:data_overview}
\begin{tabular}{lrrrr}
\toprule
 & \multicolumn{2}{c}{LBM} & \multicolumn{2}{c}{VLA Foundry} \\
\cmidrule(lr){2-3} \cmidrule(lr){4-5}
Split & Tasks & Episodes & Tasks & Episodes \\
\midrule
Real & 362 & 46,063 & 361 & 47,068 \\
Sim & 41 & 7,348 & 42 & 7,548 \\
\midrule
Total & 403 & 53,411 & 403 & 54,616 \\
\bottomrule
\end{tabular}
\end{table}

\begin{table}[htbp]
\centering
\caption{Simulation evaluation tasks. Seen tasks are used in multitask training. Unseen tasks are held out.}
\label{tab:data_per_task_breakdown}
\begin{tabular}{rlrr}
\toprule
 & & \multicolumn{2}{c}{Episodes} \\
\cmidrule(lr){3-4}
\# & Task & LBM & VLA Foundry \\
\midrule
\multicolumn{4}{l}{\textit{Seen tasks}} \\
\midrule
1 & BimanualPlaceAppleFromBowlIntoBin & 196 & 200 \\
2 & BimanualPlaceFruitFromBowlIntoBin & 196 & 200 \\
3 & BimanualPutRedBellPepperInBin & 196 & 200 \\
4 & BimanualPutSpatulaOnPlateFromDryingRack & 196 & 200 \\
5 & BimanualPutSpatulaOnPlateFromTable & 196 & 200 \\
6 & BimanualStackPlatesOnTableFromDryingRack & 196 & 200 \\
7 & BimanualStoreCerealBoxUnderShelf & 196 & 200 \\
8 & PlaceCupByCoaster & 196 & 200 \\
9 & PushCoasterToCenterOfTable & 196 & 200 \\
10 & PushCoasterToMug & 196 & 200 \\
11 & PutBananaOnSaucer & 49 & 50 \\
12 & PutKiwiInCenterOfTable & 49 & 50 \\
13 & PutMugOnSaucer & 196 & 200 \\
14 & PutSpatulaInUtensilCrock & 196 & 200 \\
15 & TurnCupUpsideDown & 490 & 500 \\
16 & TurnMugRightsideUp & 490 & 500 \\
\midrule
\multicolumn{4}{l}{\textit{Unseen tasks}} \\
\midrule
17 & BimanualPlaceAvocadoFromBowlIntoBin & 196 & 375 \\
18 & BimanualPutSpatulaOnPlateFromUtensilCrock & 195 & 400 \\
19 & PutMugInCenterOfTable & 294 & 300 \\
\bottomrule
\end{tabular}
\end{table}

\begin{table}[htbp]
\centering
\caption{Training data samples in VLA Foundry.}
\label{tab:data_manifest}
\begin{tabular}{lrrr}
\toprule
Split & Tasks & Episodes & Training samples \\
\midrule
Real & 361 & 47,068 & 17,156,497 \\
Sim & 42 & 7,548 & 1,647,049 \\
\midrule
Total & 403 & 54,616 & 18,803,546 \\
\bottomrule
\end{tabular}
\end{table}

\section{Additional Simulation Evaluation Analysis}
\label{sec:additional_sim_results}
In this section, we provide additional results in from the simulation evaluation. 

\subsection{Comparison of OS and CS Variants of LBM Eval}
Due to code changes between the paper submission and the final open-sourcing of the simulation benchmark, the evaluation results may differ slightly from \cite{lbmtri2025}. To provide context, we show the evaluation results here compared to evaluating the models on the original (closed source) simulation benchmark. We observe that the vast majority of the simulation training demonstrations were collected using a version of the simulation much closer to \lbmevalcs{}.

\subsection{Comparison of \foundryVLA{} and \foundryQwenVLA{}}
\label{sec:sim-real-both}

\begin{figure}
    \centering
    \includegraphics[width=1\linewidth]{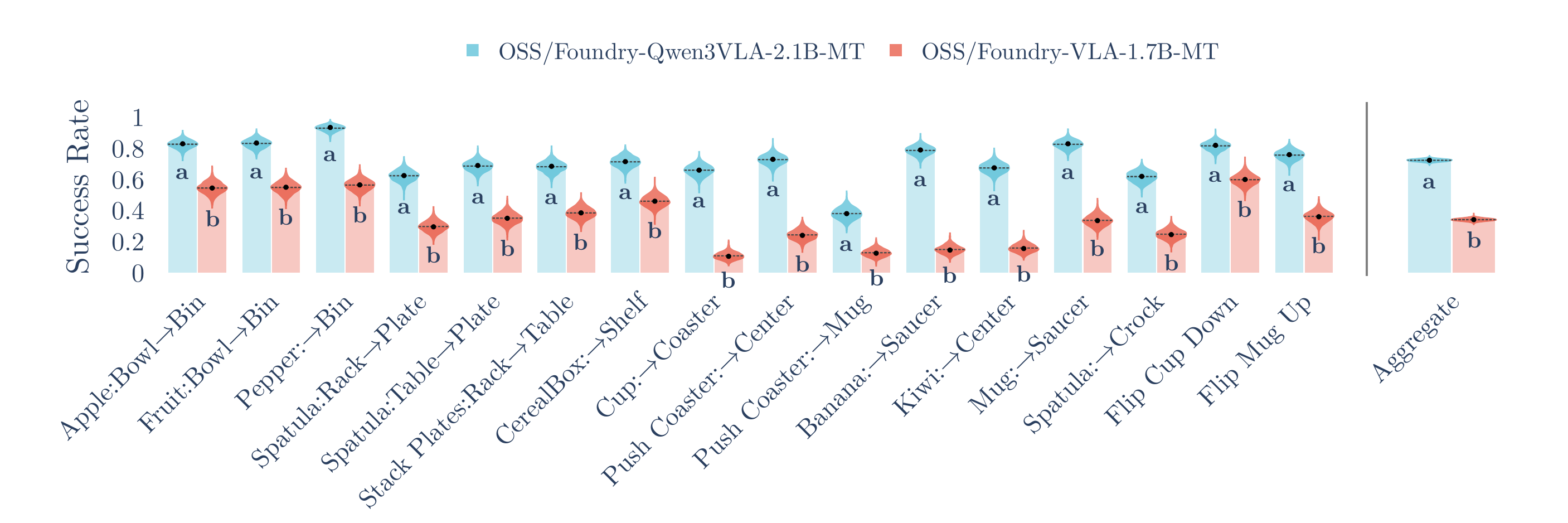}
    \caption{Comparison of \foundryQwenVLA{} and \foundryVLAMT{} models (seen tasks). The \foundryQwenVLA{} out performs than \foundryVLA{} in aggregate over the seen tasks.}
    \label{fig:foundry_qwen_vla_comparison}
\end{figure}

We also provide direct comparisons of \foundryQwenVLA{} and \foundryVLA{} models in Figure \ref{fig:foundry_qwen_vla_comparison} and \ref{fig:foundry_qwen_vla_comparison_unseen}.

\subsection{Comparison of \foundryVLA{} and \foundryQwenVLA{}}
\begin{figure}[h]
    \centering
    \begin{subfigure}[b]{0.48\textwidth}
        \centering
        \includegraphics[width=\textwidth]{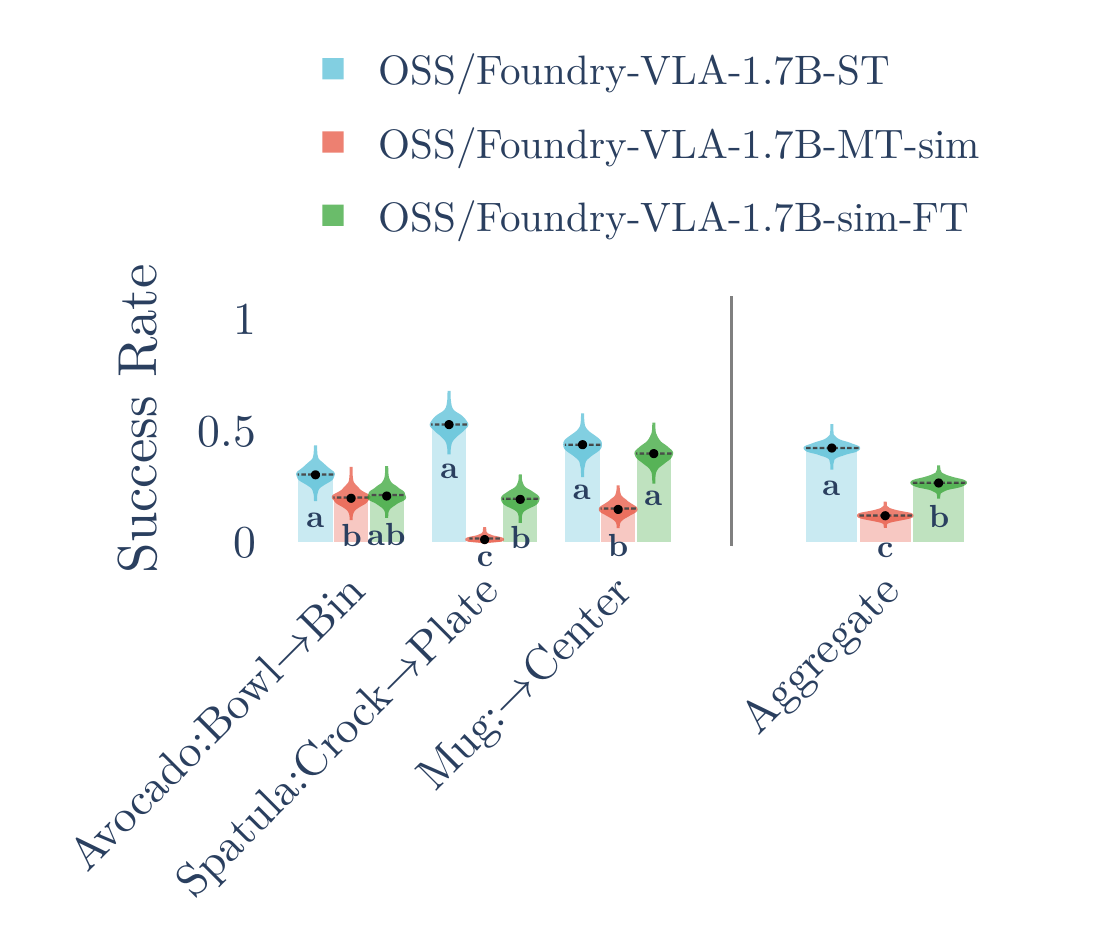}
        \caption{\foundryVLASim{} performance on unseen tasks}
        \label{fig:foundry_vla_unseen2}
    \end{subfigure}%
    \hfill
    \begin{subfigure}[b]{0.42\textwidth}
        \centering
        \includegraphics[width=\textwidth]{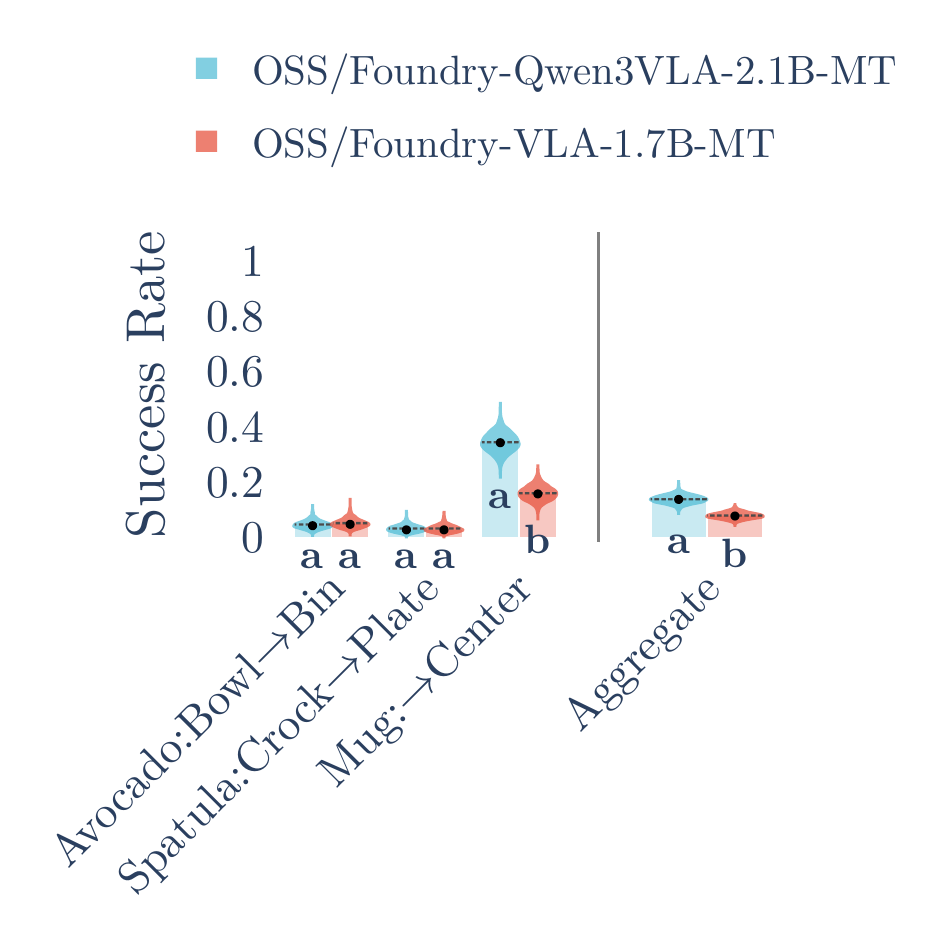}
        \caption{Comparison of \foundryQwenVLA{} and \foundryVLAMT{} on unseen tasks.}
        \label{fig:foundry_qwen_vla_comparison_unseen}
    \end{subfigure}
    \caption{Simulation results on \lbmevaloss{} (unseen tasks). All models demonstrate some non-zero success rates 0-shot.}
    \label{fig:foundry_qwen3vl2b_unseen2}
\end{figure}

Figure \ref{fig:foundry_vla_unseen2} shows the performance of the sim-only variant of the VLA Foundry model \foundryVLASim{} on unseen tasks in \lbmevaloss{}.

\section{Additional Qualitative Simulation Figures}
\label{sec:qualitative_sim}
\begin{figure}[h]
    \centering
    \includegraphics[width=\textwidth]{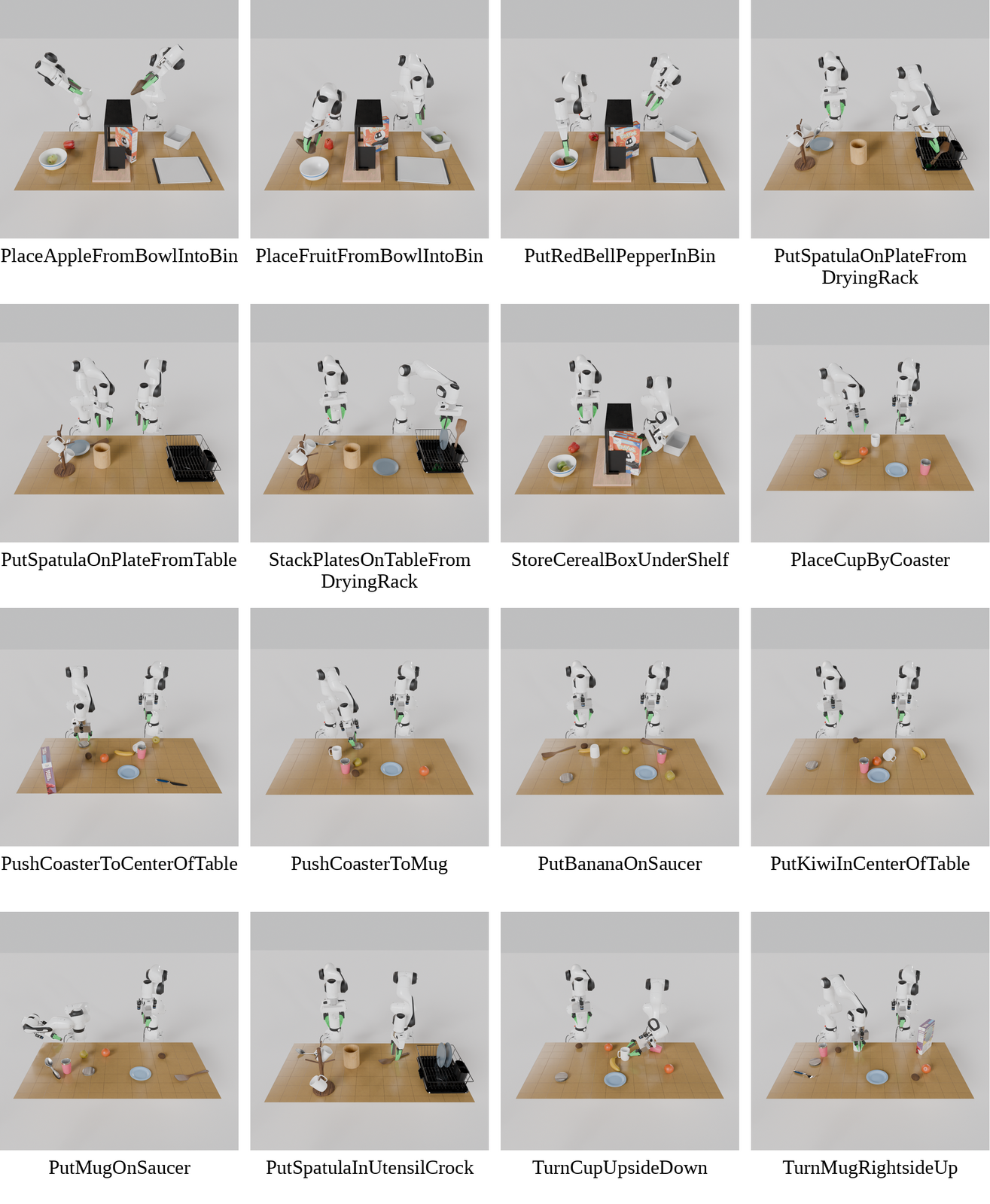}
    \caption{Overview of seen simulation evaluation tasks (failures). Here, we show a single still from about the midpoint of a failed rollout from \foundryQwenVLA{}. Videos of selected successful and failed rollouts can be found at \foundryWebsite{}.
    Companion plot to Figure \ref{fig:simulation_task_overview_successes}.}
    \label{fig:simulation_task_overview_failures}
\end{figure}

\begin{figure}[h]
    \centering
    \includegraphics[width=\textwidth]{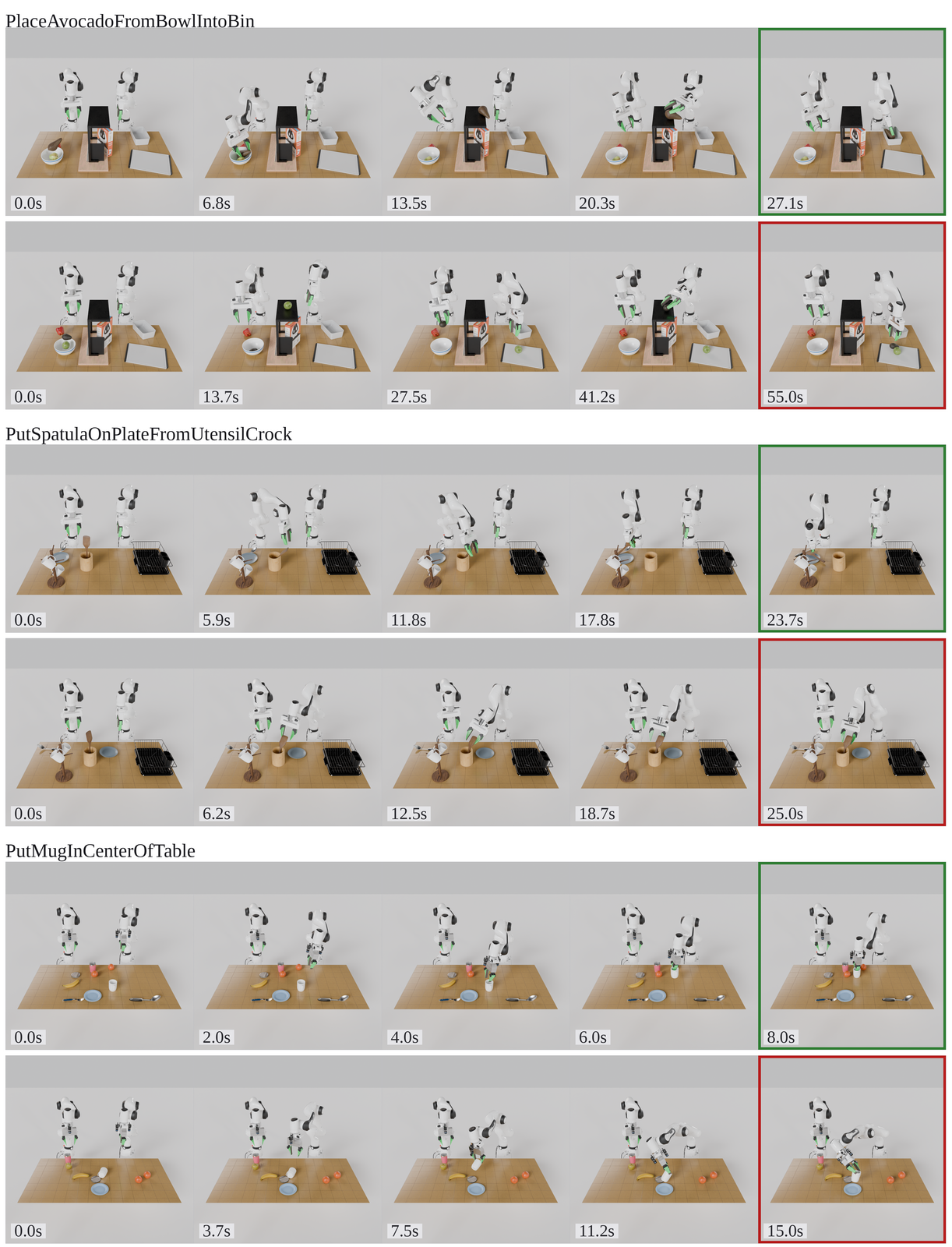}
    \caption{Example of success and failure rollouts for \foundryQwenVLA{} on tasks unseen at training time. For each task, the top row is a success and the bottom row is a failure. The timeout for each task depends on benchmark definitions of \lbmevaloss{}.}
    \label{fig:filmstrip_rollouts}
\end{figure}

\begin{figure}[htbp]
    \centering
    \includegraphics[width=0.9\textwidth]{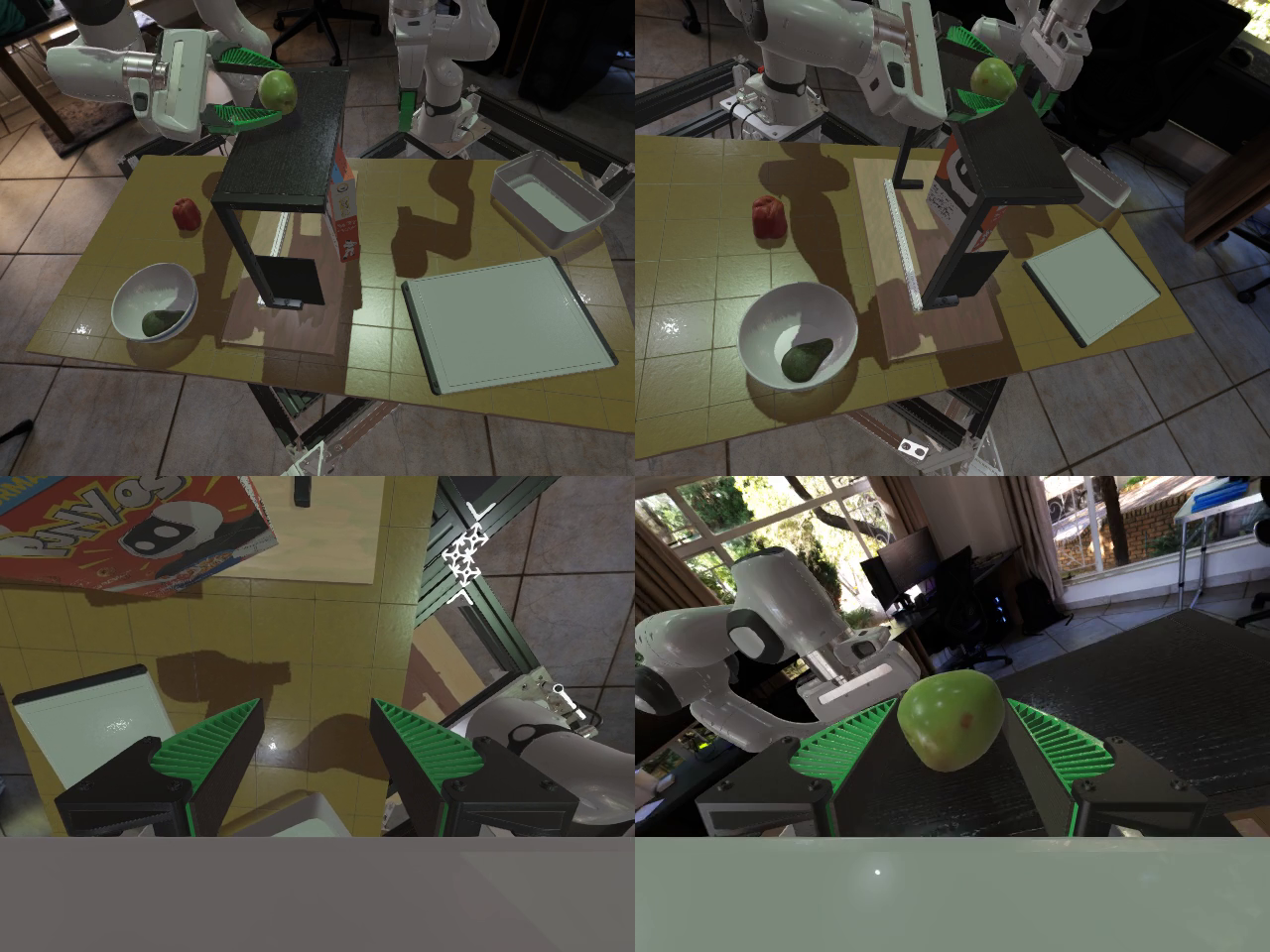}
    \caption{Example of sensor measurements at inference time. Image captured at approximately the same timestamp as the PlaceAppleFromBowlIntoBin render in Figure \ref{fig:simulation_task_overview_successes}. The images are then post processed further for input to the VLA models such as \foundryQwenVLA{} and \foundryVLA{}. While some simulation stations include an extra wrist camera per arm, \foundryQwenVLA{} and \foundryVLA{} use only the four shared cameras for VLA training and inference. Refer to \cite{lbmtri2025} for further details on the simulation stations.}
    \label{fig:raw_sensor_measurements}
\end{figure}

Figure \ref{fig:simulation_task_overview_failures} provides example snapshots of randomly sampled failure episodes from the \foundryQwenVLA{} checkpoint as a companion to Figure \ref{fig:simulation_task_overview_successes}. Figure \ref{fig:raw_sensor_measurements} gives an example of raw sensor measurements from \lbmevaloss{}. Figure \ref{fig:filmstrip_rollouts} shows temporal examples of successful and non-successful rollouts for qualitative purposes.

\end{document}